\pdfoutput=1
\documentclass[10pt,logo,copyright]{nvidiatechreport}
\usepackage[authoryear,sort&compress,round]{natbib}

\usepackage[utf8]{inputenc}
\usepackage[T1]{fontenc}

\usepackage{parskip}
\usepackage{url}
\usepackage{hyperref}
\usepackage{booktabs}
\usepackage{amsfonts}
\usepackage{nicefrac}
\usepackage{microtype}
\usepackage{xcolor}
\usepackage[dvipsnames]{xcolor}
\usepackage{graphicx}
\usepackage{tabularx}
\usepackage{makecell}
\usepackage{float}
\usepackage[section]{placeins}
\usepackage{wrapfig}

\usepackage{amsmath,amsfonts,bm,bbm}
\usepackage{multirow}

\definecolor{MyRed}{RGB}{216,56,58}
\definecolor{MyBlue}{RGB}{69,127,228}
\definecolor{MyOrange}{RGB}{238,130,47}
\definecolor{MyGreen}{RGB}{90,194,149}
\definecolor{lightblue}{rgb}{0.46,0.73,0.00}

\title{LocateAnything: Fast and High-Quality Vision-Language Grounding with Parallel Box Decoding}

\author{Shihao Wang$^{1*}$, ~~Shilong Liu$^{2*}$, ~~Yuanguo Kuang$^{1}$, ~~Xinyu Wei$^{1}$, ~~Yangzhou Liu$^{3}$, ~~Zhiqi Li, ~~Yunze Man$^{4*}$, ~~Guo Chen$^{3*}$, ~~Andrew Tao, ~~Guilin Liu, ~~Jan Kautz, ~~Lei Zhang$^{1}$, ~~Zhiding Yu$^\dagger$}

\begin{document}

\maketitle

\vspace{-8mm}

\noindent
\textbf{Links:}
{
\hypersetup{urlcolor=nvidiagreen}
\href{https://github.com/NVlabs/Eagle/tree/main/Embodied}{GitHub} |
\href{https://huggingface.co/nvidia/LocateAnything-3B}{HF Model} | 
\href{https://huggingface.co/spaces/nvidia/LocateAnything}{HF Demo} |
\href{https://research.nvidia.com/labs/lpr/locate-anything/}{Project Page}
}

\begin{figure}[h!]
    \centering
    \includegraphics[width=\linewidth]{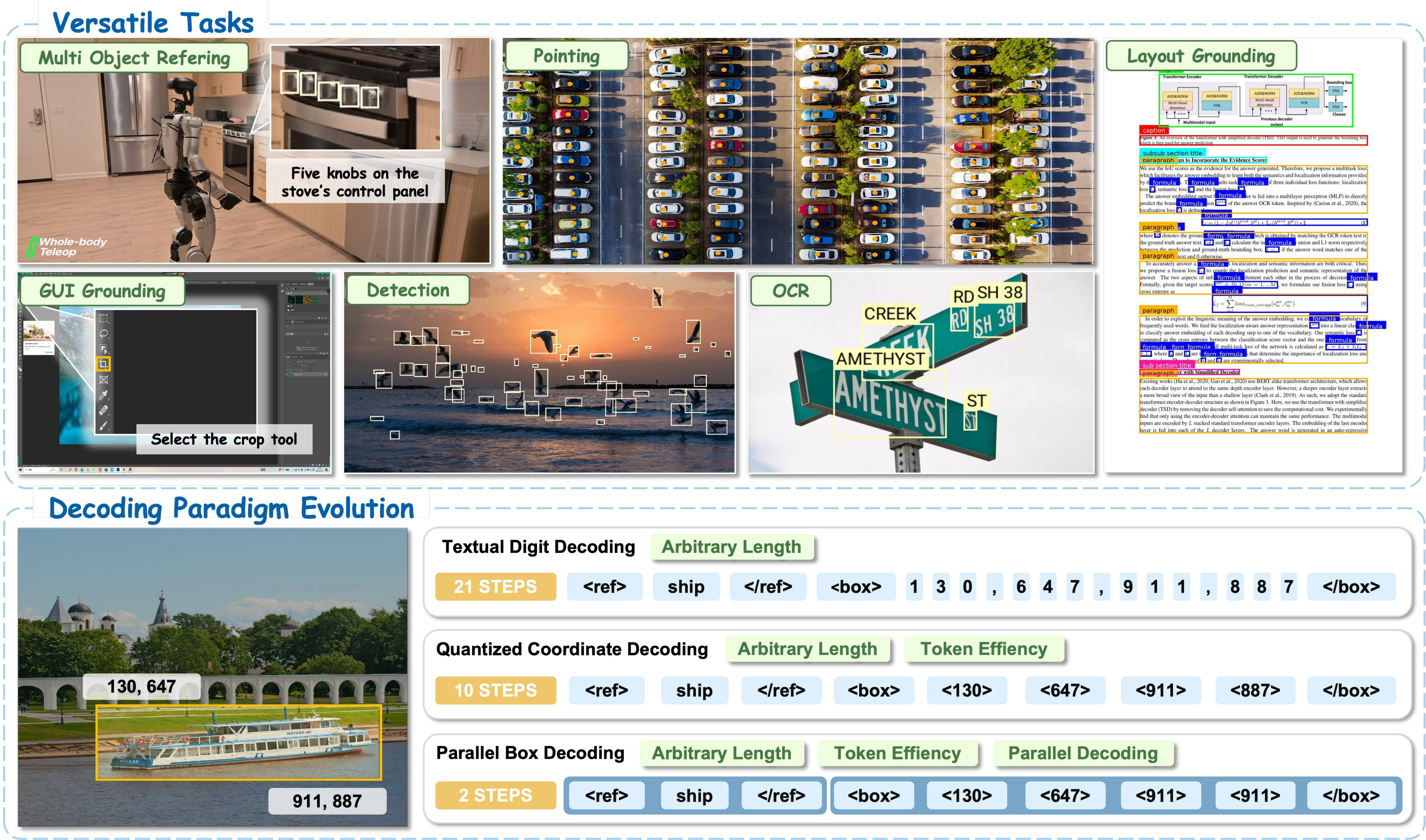} 
    \caption{\textbf{Versatile tasks of LocateAnything with parallel box decoding.}
    \textbf{Top:} LocateAnything supports diverse localization tasks under a unified vision-language model.
    \textbf{Bottom:} Textual digit decoding spells coordinates digit by digit, and quantized coordinate decoding predicts coordinate tokens sequentially.
    In contrast, \textbf{Parallel Box Decoding} predicts each geometric unit (\eg, a bounding box) in a single forward pass.}
    \label{fig:teaser}
\end{figure}

\begin{abstract}
Vision-language models (VLMs) commonly formulate visual grounding and detection as a coordinate-token generation problem, serializing each 2D box into multiple 1D tokens that are learned and decoded largely independently. This token-by-token decoding mismatches the coupled structure of box geometry and creates a practical inference bottleneck due to strictly sequential generation. We introduce \textbf{LocateAnything}, a unified generative grounding and detection framework based on \textbf{Parallel Box Decoding (PBD)}. By decoding geometric elements such as bounding boxes and points as atomic units in a single step, LocateAnything preserves intra-box geometric coherence and unlocks substantial parallelism. We show that PBD improves both decoding throughput and localization accuracy. We further develop a scalable data engine and curate \textbf{LocateAnything-Data}, a large-scale dataset with more than 138 million training samples, substantially increasing data diversity for high-precision localization. Extensive evaluations show that LocateAnything advances the speed–accuracy frontier, achieving significantly higher decoding throughput while improving high-IoU localization quality across diverse benchmarks. The results highlight the complementary benefits of Parallel Box Decoding and large-scale training data in enabling efficient and precise unified visual grounding and detection.
\vspace{-1mm}
\end{abstract}

\abscontent
\vspace{-10mm}
\section{Introduction}
Vision-language models (VLMs)~\citep{bai2025qwen2.5vl, chen2025eagle, wang2025internvl3, huang2026step3, yang2025kwai, deshmukh2025nvidia} are increasingly adopted as a general-purpose backbone for interactive and embodied systems due to their broader knowledge and stronger instruction-following capabilities than conventional specialized models \citep{zhang2022dino, liu2023grounding, carion2020end, ren2016faster}. To act in the world, VLMs \citep{bai2025qwen2.5vl, fu2025llmdet, zhan2024griffon, wang2025internvl3, azzolini2025cosmos} must be tightly grounded in \emph{perception} — in particular, they \emph{localize} task-relevant entities (\eg, objects~\citep{zhang2024llava, jiang2025rexomni, yu2025perception, wang2023exploring}, UI elements~\citep{liu2025scalecua, lin2024showui, feizi2025grounding, nayak2025ui}, regions~\citep{ren2024pixellm, yuan2025pixelrefer, lai2024lisa, cheng2024spatialrgpt, ranzinger2024radio, heinrich2025radiov2}) from natural-language intents with high quality and low latency, which requires high vision-language grounding capabilities. 

Object detection and grounding in VLMs~\citep{zhan2024griffon, li2025lmmdet, yu2025perception, peng2023kosmos, zhang2024ferretv2, jiang2025rexomni, man2025locateanything3d} are often formulated as a \emph{generative} problem. Under the next-token prediction (NTP) paradigm~\citep{chen2021pix2seq, jiang2025rexomni, peng2023kosmos}, a VLM can answer open-ended queries by emitting spatial coordinates as a token sequence. As illustrated in the bottom panel of Fig.~\ref{fig:teaser}, existing methods~\citep{you2023ferret, peng2023kosmos, zhang2024ferretv2, jiang2025rexomni, qi2025cot4det} commonly represent coordinates as either \textbf{Textual Digits} (\eg, ``1024'' as ``1'', ``0'', ``2'', ``4'') or \textbf{Quantized Tokens} (\eg, \(x_1 \rightarrow y_1 \rightarrow x_2 \rightarrow y_2\)). Despite their differences, these representations serialize a 2D geometric object into a 1D stream, forcing token-by-token generation at inference time. This token-level sequential decoding becomes a practical bottleneck (higher latency and lower throughput) and under-utilizes the strong 
structured correlation among coordinates \((x_1, y_1, x_2, y_2)\).

\begin{figure}[t]
\centering
\includegraphics[width=\linewidth]{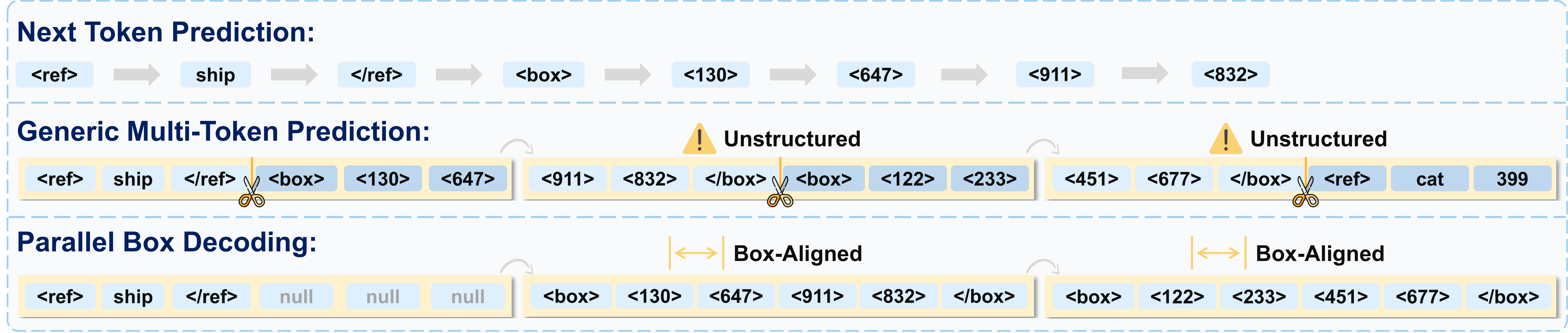}
\caption{\textbf{Comparison of Token Decoding Methods.} The NTP generates coordinate values one-by-one. The standard MTP method results in irregular distributions and non-coherent, unstructured patterns. Our proposed PBD generates a single atomic box (or point) unit in a parallel step, ensuring box-aligned and structured output.}
\label{fig:method_comparison}
\end{figure}

Multi-Token Prediction (MTP)~\citep{li2025diffusionvl, liu2025sequential, nie2025large, ye2025dream} offers a natural approach to reducing decoding steps by predicting multiple tokens in parallel. In language modeling, MTP is usually implemented by randomly (i) choosing positions in the sequence and training the model to predict a following span in parallel (\ie, next-block prediction)~\citep{liu2025sequential, cai2024medusa, li2025eagle, liu2024deepseek}, or (ii) masking some tokens of the sequence and training the model to reconstruct the original text, such as masked diffusion modeling~\citep{li2022diffusion, arriola2025block, nie2025large, liu2025tidar}. However, these formulations are largely \emph{structure-agnostic}: they treat inputs as generic token streams and mainly capture correlations driven by co-occurrence. Inferring the missing tokens from random subsets requires the model to represent complex and irregular conditional distributions. For tightly coupled units such as bounding boxes, this supervision does not match well the training objective because it can learn to generate token combinations across bounding-box boundaries and even object categories, as demonstrated in Fig.~\ref{fig:method_comparison}. Consequently, the model must fit many unreliable patterns, inducing spurious correlations, sacrificing structured decoding, and amplifying error propagation, which together reduce accuracy, reliability, and decoding speed.

To reconcile high-throughput decoding with reliable localization, we propose \textbf{LocateAnything}, a unified framework for VLM-based visual detection and grounding built upon \textbf{Parallel Box Decoding (PBD)}. Our key idea is to align MTP blocks with structured units: during training, LocateAnything treats each bounding box (or point) as an \emph{atomic unit} and learns to predict the full coordinate set \((x_1, y_1, x_2, y_2)\) in one parallel step. This \emph{box-aligned} training target avoids arbitrary chunking of coordinate tokens. As a result, our strategy improves the localization performance of the model, while simultaneously unlocking the speed benefits of parallel decoding.

With the proposed PBD, we study various strategies for structured bounding-box decoding to balance throughput and accuracy. Our observations motivate a  flexible inference design to meet different latency--robustness requirements by providing three on-demand modes. (i) \textbf{Fast Mode} (MTP) predicts full boxes in parallel for maximum throughput, which is suitable for latency- and compute-constrained settings, such as on-device robotics and embodied agents. (ii) \textbf{Slow Mode} (NTP) decodes coordinate tokens autoregressively for maximum stability, which is appropriate for high-precision labeling, final-pass dataset curation, and accuracy-oriented offline evaluation. (iii) \textbf{Hybrid Mode} uses Fast Mode by default and falls back to Slow Mode when the parallel output is unreliable, \eg, due to format or consistency violations; this mode is intended for production pipelines that require both speed and accuracy. Overall, Hybrid Mode preserves most of the speed gains of parallel decoding while maintaining robust outputs.

\vspace{+1mm}
Our main contributions are summarized as follows:
\vspace{-2mm}
\begin{itemize}
\item We introduce \textbf{LocateAnything}, an early exploration of applying multi-token prediction to VLM-based detection/grounding via \textbf{Parallel Box Decoding}, performing box-aligned decoding to improve throughput and accuracy.
\item We present a Hybrid decoding policy that detects unreliable parallel blocks and performs localized NTP re-decoding only for the problematic block, reducing worst-case failures while retaining most speed gains.
\item Extensive evaluations, including layout grounding, long-tail detection, and GUI grounding, show that LocateAnything advances the \textbf{speed--accuracy frontier}, outperforming the SOTA by a large margin. It achieves up to \textbf{2.5$\times$} higher decoding throughput while improving localization quality.
\end{itemize}
\vspace{-2mm}

\section{Related Work} \label{sec:related}

\textbf{Visual Detection and Grounding in VLMs}. Visual grounding/detection tasks traditionally rely on task-specific heads~\citep{carion2020end, liu2024grounding, ren2016faster, jiang2024far3d}, but recent VLMs like Qwen-VL series~\citep{bai2025qwen2.5vl, bai2025qwen3vltechnicalreport}, InternVL~\citep{chen2024internvl} and Shikra~\citep{chen2023shikra} formulate it as an autoregressive token generation problem. This generative paradigm, however, often suffers from structural hallucinations and high latency~\citep{li2023evaluating}. To mitigate these issues, Rex-Omni~\citep{jiang2025rexomni} employs point-based prediction, while Patch-as-Decodable-Token (PaDT)~\citep{su2026patch} and Groma~\citep{ma2024groma} utilize visual reference tokens to point directly to image patches. Complementary innovations such as Pink~\citep{xuan2024pink}, ViP-LLaVA~\citep{cai2024vipllava}, Griffon~\citep{zhan2024griffon},  DnU~\citep{lin2024draw} and PAM~\citep{lin2025perceive} focus on enhancing 2D referential comprehension through visual prompt engineering and multi-granularity feature scaling.
LLMDet~\citep{fu2025llmdet} boosts detection recall by data distribution tuning. To bypass serial decoding bottlenecks, WeDetect~\citep{fu2025wedetect} treats detection as a parallel retrieval task. Advanced perception logic is further integrated via Chain-of-Thought (CoT)~\citep{qi2025cot4det}, while post-training strategies such as Vision-R1~\citep{zhan2025visionr1}, UniVG-R1~\citep{bai2025univg} and GW-VLM~\citep{jiang2026gwvlm} utilize reinforcement learning to align model outputs with visual feedback and reduce grounding errors~\citep{zhang2024ferretv2}.

\noindent\textbf{Parallel Decoding via MTP and Diffusion LLMs}.
To mitigate autoregressive latency, parallel generation techniques such as MTP~\citep{gloeckle2024better, cai2024medusa, samragh2025your} predict multiple future tokens simultaneously, often coupled with speculative decoding to accelerate inference. Recent extensions such as Future Summary Prediction~\citep{mahajan2025beyond} capture long-term dependencies via auxiliary heads. Concurrently, Diffusion Language Models (DLMs) such as LLaDA~\citep{nie2025large}, Dream~\citep{ye2025dream}, and DiffuCoder~\citep{gong2025diffucoder} frame sequence generation as a discrete denoising process, enabling bidirectional context modeling and non-autoregressive decoding. Hybrid semi-autoregressive paradigms, including Block Diffusion~\citep{arriola2025block}, SDLM~\citep{liu2025sequential} and Fast-dLLM v2~\citep{wu2025fast}, decode fixed-size token blocks in parallel while maintaining causal dependencies to preserve KV-caching compatibility. More advanced frameworks~\citep{wang2025diffusion, lu2025adablock} unlock inter-block parallelism and adaptive block scheduling. These paradigms have been extended to the multimodal domain via DiffusionVL~\citep{li2025diffusionvl}, translating autoregressive LMMs into high-performance diffusion-based models.

\textbf{LocateAnything} differs from existing works in two key aspects. First, instead of generating bounding boxes via slow NTP, we output the complete box in a single parallel step.
Second, recent MTP paradigms group tokens into arbitrary chunks.
Instead, our PBD treats the entire coordinate set as a single atomic block, resolving both the fragmentation of NTP and the arbitrary chunking of MTP, seamlessly unifying high throughput with structural coherence.

\begin{figure}[t]
\centering
    \includegraphics[width=\linewidth]{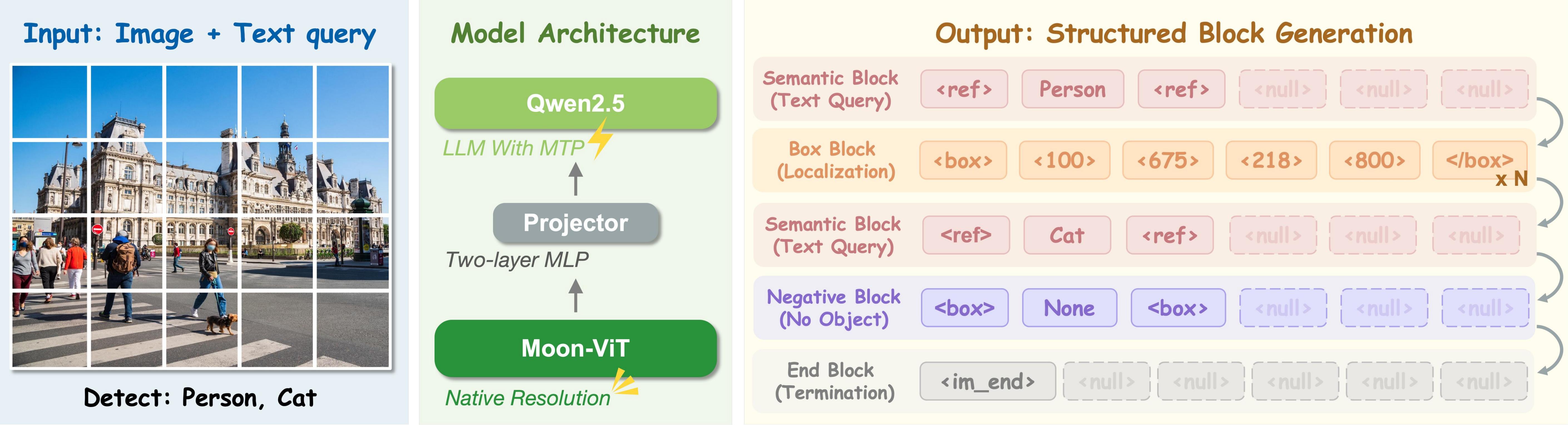} 
    \caption{\textbf{Architecture and Block-Based Output Representation.} LocateAnything formulates localization as generating a sequence of fixed-length, \emph{box-aligned atomic blocks}. Four functional block types—Semantic, Box, Negative, and End blocks—are defined to jointly specify predicted entities or termination states.}
    \label{fig:method1}
\end{figure}
\section{Method}

This section presents \textbf{LocateAnything}, a fast and effective framework that integrates \textbf{Parallel Box Decoding (PBD)} into VLMs for visual detection and grounding. Section~3.1 introduces the model architecture and the block-based output formulation. Section~3.2 details the joint training strategy, which aligns NTP with block-level MTP. Section~3.3 describes the on-demand inference mechanism, featuring a hybrid mode that dynamically balances decoding throughput and robustness. Finally, Section~3.4 outlines the construction of our large-scale training dataset, \textbf{LocateAnything-Data}.

\vspace{-2mm}
\subsection{Model Architecture and Formulation}

\noindent\textbf{Overview.}
As illustrated in Fig.~\ref{fig:method1}, LocateAnything builds upon a native-resolution VLM pre-trained on large-scale image-text corpora. The architecture comprises a Moon-ViT~\citep{team2025kimi} vision encoder and a Qwen2.5~\citep{qwen2.5} language decoder, bridged by a MLP projector. Given an input image \(\mathcal{I}\), the vision encoder extracts visual tokens \(Z = \text{Encoder}(\mathcal{I})\) at the native resolution, preserving the fine-grained spatial details crucial for high-precision localization. These tokens are subsequently fed into the language model, which directly converts them into a sequence of box-aligned block-level predictions.

\noindent\textbf{Block-Based Output Formulation.}
To facilitate PBD, we abandon standard NTP coordinate generation. Instead, continuous coordinates are normalized to $[0, 1000]$, discretized into tokens~\citep{jiang2025rexomni, chen2021pix2seq}, and reorganized into a sequence of blocks $\mathbf{B} = (b_1, b_2, \dots, b_N)$. Conditioned on the visual features $Z$ and a text query $\mathcal{E}$, the joint probability is formulated as $P(\mathbf{B} \mid \mathcal{Z}, \mathcal{E}) = \prod_{i=1}^{N} P(b_i \mid b_{<i}, Z, \mathcal{E})$.

Each block \(b_i\) acts as an atomic unit of constant length \(L=6\), accommodating a bounding box and two structural tokens (\eg, \texttt{<box>} and \texttt{</box>}). To guarantee uniform tensor shapes for parallel decoding, any unoccupied positions are padded with a \texttt{<null>} token. As depicted in Fig.~\ref{fig:method1}, we define four functional block types. \textit{(1) Semantic Block}: Encodes the linguistic identity. If an expression exceeds the capacity of a single block, it is partitioned across multiple consecutive blocks.
\textit{(2) Box Block}: Uses four quantized coordinates representing the bounding boxes.
\textit{(3) Negative Block}: Explicitly indicates the absence of a queried object.
\textit{(4) End Block}: Signals the termination of the generation process.

\subsection{Training Design}
Our method treats bounding box coordinates as an indivisible atomic unit, enforcing structured supervision and unlocking the capability for parallel generation. However, parallelizing the output directly in the training phase risks disrupting the model's inherent causal reasoning process. To resolve this issue, we introduce a dual-formulation training strategy that jointly optimizes two aligned representations: the NTP sequence to preserve the causal reasoning ability, and the block-wise MTP formulation for box-aligned predictions. To implement this, a single concatenated input sequence is constructed: $x_{\text{all}} = x_{\text{vis}} \oplus x_{\text{q}} \oplus x_{\text{ntp}} \oplus x_{\text{blk}}$, where \(\oplus\) denotes sequence concatenation. The terms \(x_{\text{vis}}\) and \(x_{\text{q}}\) serve as the shared context (visual and text query inputs), \(x_{\text{ntp}}\) represents the standard NTP input sequence, and \(x_{\text{blk}}\) is the block-wise MTP input sequence. Essentially, they represent the identical ground truth in two distinct formats: a \textit{token-level representation} and a \textit{block-level representation}.

Specifically, inspired by~\citep{liu2025sequential, liu2025wedlm}, \(x_{\text{blk}}\) is constructed by traversing \(x_{\text{ntp}}\) from left to right, splitting and padding the sequence according to our previously defined block rules. Within each block, we retain the first token to serve as the prediction context, while replacing all subsequent tokens with \texttt{[mask]} tokens. This structure prompts the model to simultaneously predict all masked tokens within the block in a single cohesive step. Notably, if the block size is set to 1, this MTP formulation naturally becomes equivalent to standard NTP.

\vspace{+1mm}
\begin{wrapfigure}{r}{0.52\textwidth}
    \centering
    \vspace{-5mm}
    \includegraphics[width=0.5\textwidth]{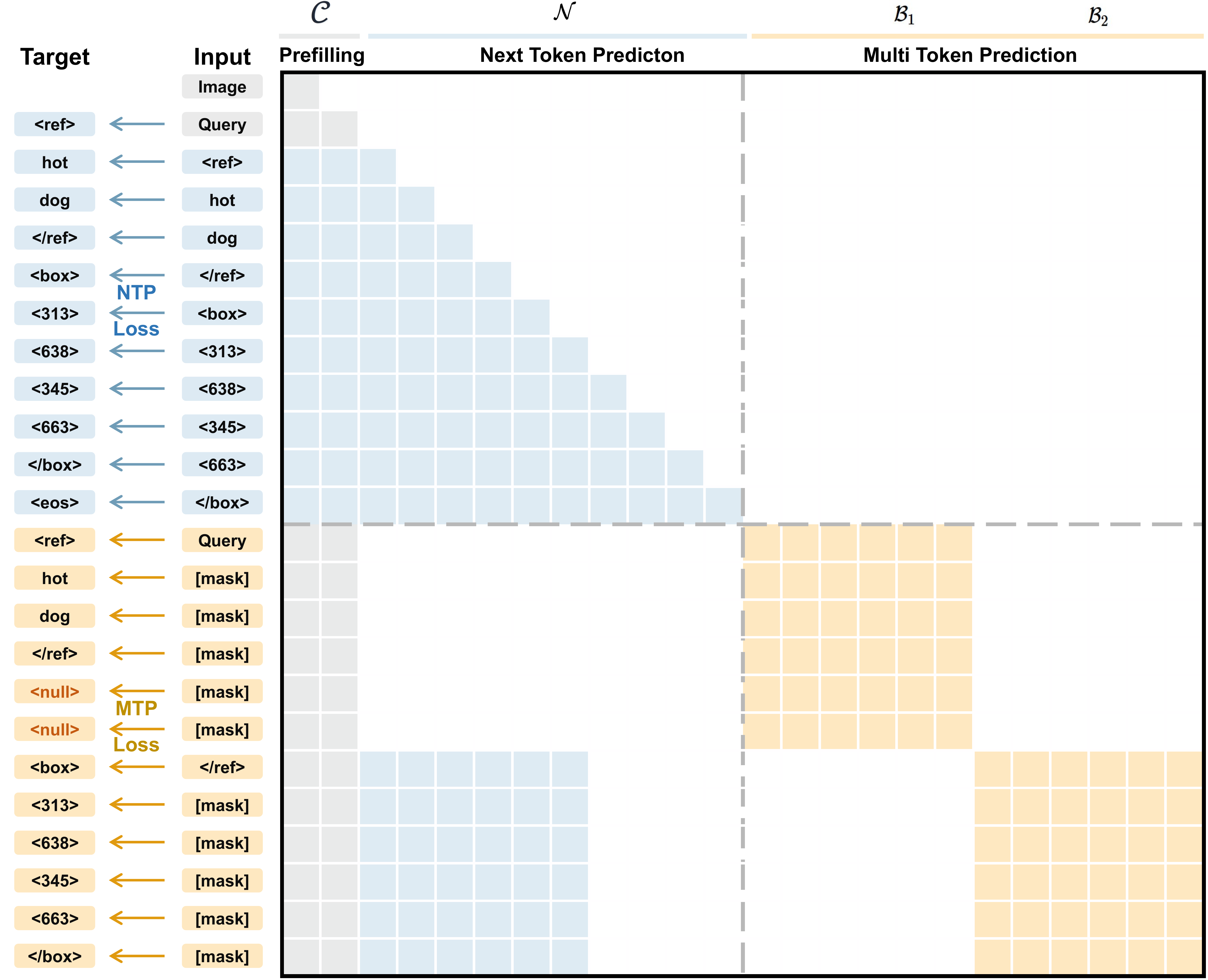}
    \vspace{-2mm}
    \caption{\textbf{Attention Mask for Joint NTP--MTP Training.} The shared context and NTP stream use causal attention, the MTP blocks follow a block-causal pattern across blocks, and tokens within the same block share bidirectional attention. The two streams are isolated to prevent leakage while jointly attending to the shared context.}
    \label{fig:attn_mask}
    \vspace{-4mm}
\end{wrapfigure}
\noindent\textbf{Attention Mask Design.} 
The core challenge of this dual-sequence formulation is how to isolate the NTP and MTP streams while allowing both to leverage the shared context. This is achieved through a specialized attention mask (as shown in Fig.~\ref{fig:attn_mask}), which dictates information flow via three distinct behaviors:
 
\noindent\textit{Causal Attention for NTP.} To preserve the original language capabilities of the VLM, the shared context (\(x_{\text{vis}}\) and \(x_{\text{q}}\)) and the NTP sequence (\(x_{\text{ntp}}\)) collectively employ a causal attention mask. Tokens within these segments can only attend to preceding tokens. Crucially, they are restricted from attending to \(x_{\text{blk}}\) to prevent data leakage. This strict causal formulation  perfectly aligns with the standard KV Cache usage during inference.

\noindent\textit{Causal Flow Across Blocks.} To align with the semi-autoregressive generation process, attention across different blocks in \(x_{\text{blk}}\) is strictly causal. Tokens in the active block can attend to the shared context and all previously committed blocks, but cannot see future blocks. This historical visibility enables the model to learn dependencies between different box predictions, effectively mitigating duplicate or missing bounding boxes.

\noindent\textit{Bidirectional Intra-Block Attention.} Following the block-causal design widely adopted in recent generative modeling~\citep{arriola2025block, nie2025large, wang2025diffusion, wu2025fast, fu2025efficient, wu2025fastv1}, tokens within the same block share bidirectional attention. This fully-connected intra-block interaction allows the model to capture complex internal relationships (\eg, geometric dependencies among a set of coordinates) and resolve all internal tokens simultaneously within a single functional unit.

\vspace{+1mm}
\noindent\textbf{Objective.} Guided by this mask, we jointly minimize the cross-entropy losses for both sequences, \ie, $\mathcal{L} = \mathcal{L}_{\mathrm{ntp}} + \mathcal{L}_{\mathrm{mtp}}$.

\begin{figure}[t]
    \centering
    \includegraphics[width=\linewidth]{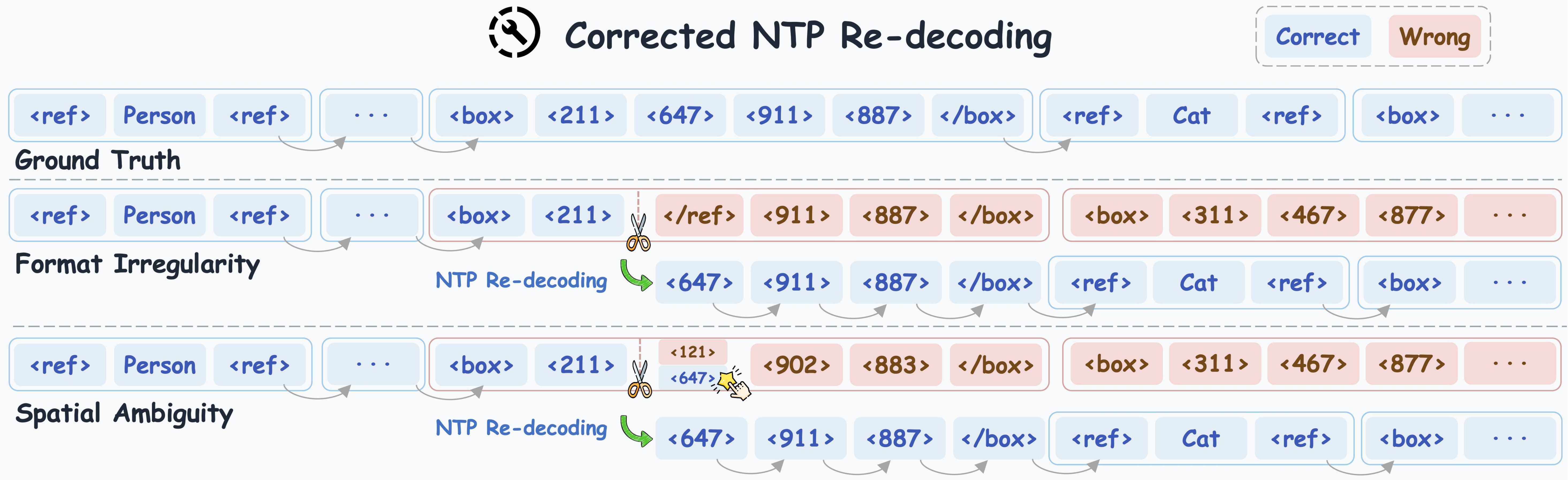} 
    \caption{\textbf{Corrected NTP Re-decoding.} When parallel decoding encounters \textit{Format Irregularity} or \textit{Spatial Ambiguity}, the model discards the erroneous block and reverts to standard NTP to ensure robust predictions.}
    \label{fig:method2}
\end{figure}

\vspace{-2mm}
\subsection{On-Demand Inference Modes}
While our proposed PBD significantly accelerates inference, parallel decoding faces an inherent exploration-exploitation dilemma in highly complex scenes, as shown in Fig.~\ref{fig:method2}. The first is \textit{Format Irregularity}, which occurs in complex scenes containing multiple instances across categories. During parallel decoding, the model may struggle at category boundaries, hesitating between continuing to predict for the current class or transitioning to a new class. This uncertainty manifests as malformed syntax within a single predicted block, erroneously mixing structural and coordinate tokens (\eg, \texttt{<box><211></ref><911><887></box>}). The second is \textit{Spatial Ambiguity}, which arises when objects are densely arranged in regular grids, such as rows or columns. The MTP approach can blur spatial boundaries and output an intermediate coordinate situated between two objects, consequently producing low IoU predictions.

Both failure patterns can be effectively resolved using an NTP fallback mechanism. The NTP prediction can achieve higher precision when handling complex category transitions and dense spatial layouts. Therefore, during MTP inference, we continuously validate the syntactic integrity and monitor spatial confidence. Specifically, an ambiguity trigger is activated if two conditions are met simultaneously: (1) the top-1 coordinate token's probability is below 0.7, and (2) the max-min difference among the top-5 coordinate tokens exceeds 80 within the [0, 1000] normalized space. Upon detection of a format violation or high spatial ambiguity, the compromised block is discarded, and the generation reverts to the last verified prefix. NTP is then employed to autoregressively generate the tokens for the specific problematic block. Once the block is completed, the model seamlessly switches back to MTP for subsequent predictions.

Based on the above discussion, we propose three on-demand inference modes to balance throughput and spatial robustness.
\textit{(1) Slow Mode}, which generates the output token-by-token using standard NTP.
\textit{(2) Fast Mode}, which leverages MTP to predict box-aligned blocks. For each block, \texttt{<null>} padding tokens are discarded, and the remaining tokens are appended to the output; the committed tokens are stored in the key-value cache and serve as causal context for subsequent prediction steps.
\textit{(3) Hybrid Mode}, which employs MTP by default but seamlessly switches to NTP when parallel outputs become unreliable.

\vspace{+1mm}
\noindent\textbf{Inference-Time Attention Mask.}
During inference, the attention mask for each MTP decoding step mirrors the training-time block-causal pattern illustrated in Fig.~\ref{fig:attn_mask}. All previously committed tokens in the KV cache follow standard causal attention, while the $n_{\text{future}}$ tokens in the current MTP block attend to each other bidirectionally, enabling parallel token prediction. Meanwhile, the current block can attend to all preceding blocks but is prevented from accessing subsequent ones. After each MTP step, the KV cache is truncated to retain only committed tokens, evicting mask tokens and the duplicated anchor to ensure the cache stays consistent with the causal prefix seen during training.

\begin{figure*}[t]
    \centering
    \includegraphics[width=\textwidth]{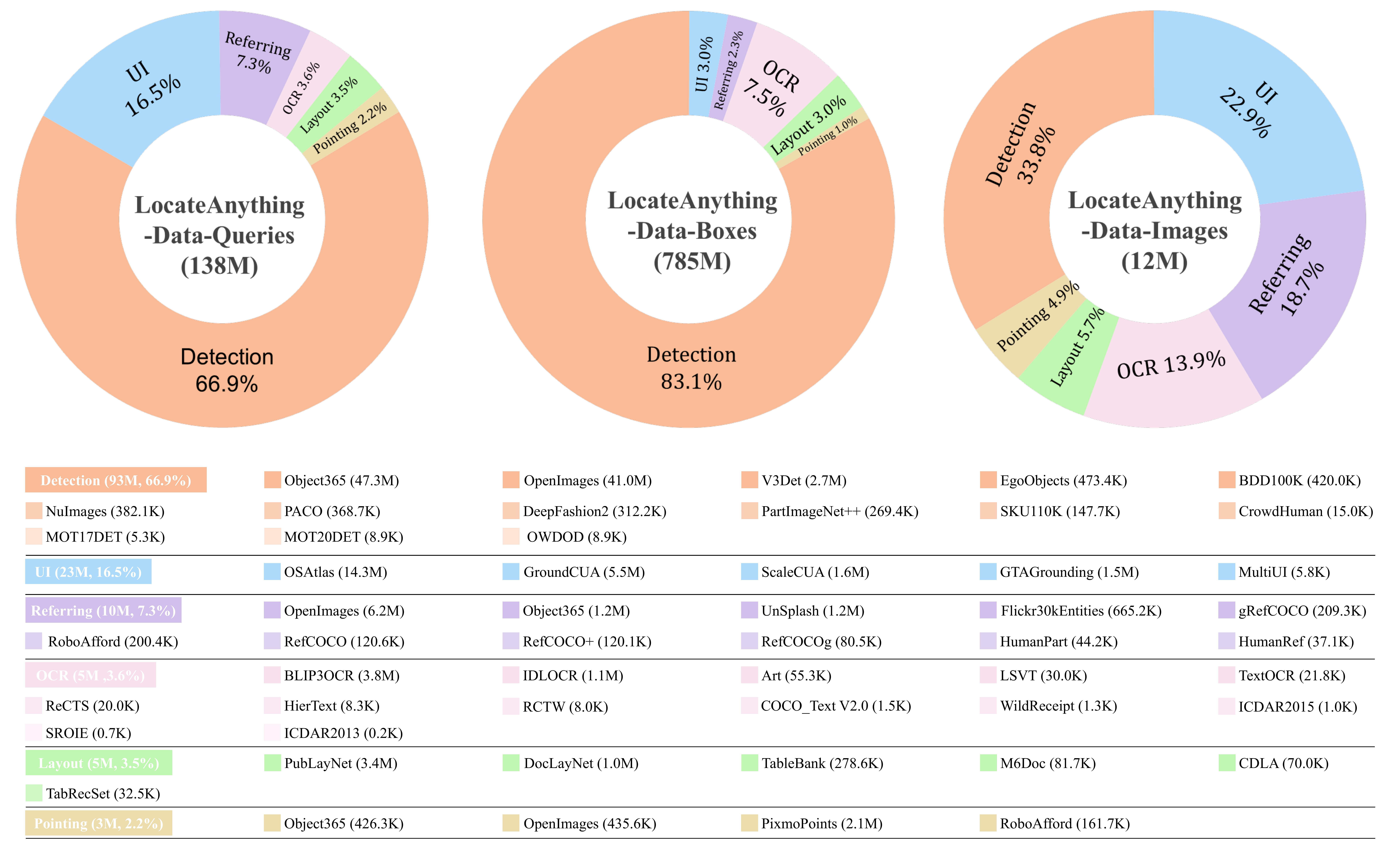}
    \caption{Overview of the \textbf{LocateAnything-Data} dataset. The pie charts illustrate the task distribution across natural language queries, bounding boxes, and unique images. The bottom panel provides a detailed breakdown specifically for the language queries, showing the absolute count and percentage for each task category.}
    \label{fig:dataset_stats}
\end{figure*}

\subsection{LocateAnything-Data}
\label{sec:dataset}

To train a highly capable model for general-purpose visual detection and grounding, we curate \textbf{LocateAnything-Data}, a large-scale, multi-domain dataset. The dataset construction details can be found in the \textbf{supplementary}. 

As illustrated in Fig.~\ref{fig:dataset_stats}, the dataset contains 12M unique images and 138M natural language queries. Furthermore, the dataset includes 785M annotated bounding boxes, providing massive and dense supervisory signals to guide the spatial learning of the LocateAnything model. The training corpus is categorized into six distinct tasks. (1) \textbf{General object detection} constitutes the foundation, representing 66.9\% of the queries and providing the essential bounding box supervision (83.1\%) to help the model achieve precise and dense coordinate alignments. (2) \textbf{Grounding user interface} elements (16.5\% of queries) enable the model to support embodied agents and graphical user interface navigation tasks. (3) \textbf{Natural language referring} comprehension (7.3\% of queries) enables the model to link complex linguistic intents to specific spatial regions. (4) \textbf{Text localization} (3.6\% of queries) ensures that the model can perceive and tightly ground textual information within images. (5) \textbf{Document and scene layout grounding} (3.5\% of queries) enriches the structural reasoning capabilities of the model. (6) \textbf{Point-based localization} tasks (2.2\% of queries) further refine the spatial precision of the model for fine-grained predictions.

\section{Experiments}
\subsection{Training Details and Evaluation Setup}
\noindent\textbf{Training Details.}
We first conduct an initial training on the base VLM with focus entirely on world-knowledge alignment, during which all detection and grounding data are excluded. We then apply a two-stage supervised fine-tuning to the base VLM to train our LocateAnything model. In Stage-1, we incorporate a massive mixture of 138M queries into the overall training data to equip the model with comprehensive grounding and detection capabilities. In Stage-2, we reduce the proportion of general training data to 20\% while significantly increasing the proportion of data containing many objects per image (\eg, MOT20Det~\citep{dendorfer2020motchallengebenchmarksinglecameramultiple}, SKU110K~\citep{goldman2019precise}) to enhance the model's ability in dense detection. For model ablations, we train all models exclusively on the COCO dataset~\citep{lin2014microsoft} to strictly isolate PBD's architectural benefits from our massive 138M data. Detailed configurations for both the base VLM and the subsequent LocateAnything model training are provided in the \textbf{supplementary materials}.
\begin{table*}[t]
\centering
\caption{Results on LVIS and COCO. Throughout all tables, ``-'' means that the information was not reported in the respective papers or the model does not support the corresponding task, \textbf{bold} and \underline{underline} highlight the best and second-best, and BPS (Boxes Per Second) measures decoding throughput.}
\vspace{-1mm}
\label{tab:lvis_coco_f1_merged}
\resizebox{1.0\textwidth}{!}{
\setlength{\tabcolsep}{5pt}
\begin{tabular}{l|c|c|ccc|c|ccc}
\toprule
\multirow{2}{*}{\textbf{Method}} &
\multirow{2}{*}{\begin{tabular}[c]{@{}c@{}}\textbf{Throughput}\\ \end{tabular}} &
\multirow{2}{*}{\begin{tabular}[c]{@{}c@{}}\textbf{Zero-Shot}\\ \textbf{(LVIS)} \end{tabular}} &
\multicolumn{3}{c|}{\textbf{LVIS} (F1@IoU)} &
\multirow{2}{*}{\begin{tabular}[c]{@{}c@{}}\textbf{Zero-Shot}\\ \textbf{(COCO)}\end{tabular}} &
\multicolumn{3}{c}{\textbf{COCO} (F1@IoU)} \\
\cline{4-6}\cline{8-10}
& & &
0.5 & 0.95 & Mean &
&
0.5 & 0.95 & Mean \\

\midrule
\multicolumn{10}{c}{{Open-set Specialized Detectors}} \\
\midrule

Grounding DINO-Swin-T~\citep{liu2023grounding} & - & Yes & 47.7 & \underline{22.7} & 38.8 & Yes & 69.8 & \underline{23.0} & \underline{56.6} \\

\midrule

\multicolumn{10}{c}{{Closed-set Specialized Detectors}} \\
\midrule
Faster RCNN-R50~\citep{ren2016faster}        & - & - & - & - & - & No & 60.6 & 7.1  & 48.1 \\
DETR-R50~\citep{carion2020end}               & - & - & - & - & - & No & 65.9 & 13.6 & 48.3 \\
Deformable-DETR-R50~\citep{zhu2021deformabledetrdeformabletransformers}    & - & - & - & - & - & No & 69.7 & 17.7 & 54.7 \\
DINO-R50~\citep{zhang2022dino}               & - & - & - & - & - & No & 68.8 & 21.1 & 55.6 \\
DINO-Swin-L~\citep{zhang2022dino}            & - & - & - & - & - & No & \textbf{75.6} & \textbf{25.4} & \textbf{62.1} \\

\midrule
\multicolumn{10}{c}{{Vision-Language Models}} \\
\midrule
DeepSeek-VL2-Small~\citep{wu2024deepseekvl2mixtureofexpertsvisionlanguagemodels}  & - & - & 56.2 & 21.0 & 41.8 & - & 60.9 & 14.9 & 45.9 \\
MiMo-VL-7B~\citep{coreteam2025mimovltechnicalreport}          & 1.0 & - & 49.5 & 8.8  & 31.4 & - & 56.5 & 6.7  & 35.9 \\
OVIS2.5-2B~\citep{lu2025ovis25technicalreport}          & 1.3 & - & 54.4 & 15.8 & 37.4 & - & 56.2 & 10.3 & 38.7 \\
Qwen3-VL-4B~\citep{bai2025qwen3vltechnicalreport}         & 1.1 & - & 59.8 & 20.0 & 43.5 & - & 63.0 & 14.2 & 46.1 \\ 
Qwen3-VL-8B~\citep{bai2025qwen3vltechnicalreport}         & 1.0 & - & 61.5 & 20.2 & 44.8 & - & 62.8 & 14.0 & 45.7 \\ 
Cosmos-Reason2-8B~\citep{cosmosreason2}         & 1.0 & - & 56.4 & 9.8 & 40.2 & - & 56.4 & 9.8 & 39.3 \\ 
SEED1.5-VL~\citep{guo2025seed15vltechnicalreport}          & - & Yes & \textbf{65.6} & 19.5 & 46.7 & Yes & 71.3 & 14.3 & 51.4 \\
Rex-Omni-3B~\citep{jiang2025rexomni}             & \underline{5.0} & Yes & \underline{64.3} & 20.7 & \underline{46.9} & Yes & \underline{72.0} & 15.9 & 52.9 \\
\rowcolor{lightblue!10}
LocateAnything-3B      & \textbf{12.7} & Yes & 62.3 & \textbf{31.1} & \textbf{50.7} & Yes & 70.1 & 19.3 & 54.7 \\
\bottomrule
\end{tabular}
}
\end{table*}

\vspace{+1mm}
\noindent\textbf{Compared Methods.} We compare LocateAnything against three categories of methods. \textbf{(1) Specialized detectors}, including representative general detection models such as DETR~\citep{carion2020end} and Deformable-DETR~\citep{zhu2021deformabledetrdeformabletransformers}, \etc, open-set detectors such as Grounding DINO~\citep{liu2023grounding}, leading document layout analysis model DocLayout-YOLO~\citep{zhao2024doclayout}, and text detection model PaddleOCRv5~\citep{cui2025paddleocr}. \textbf{(2) General-purpose VLMs with grounding capabilities}, including Qwen3-VL~\citep{bai2025qwen3vltechnicalreport}, DeepSeek-VL2~\citep{wu2024deepseekvl2mixtureofexpertsvisionlanguagemodels}, OVIS2.5~\citep{lu2025ovis25technicalreport}, MiMo-VL~\citep{coreteam2025mimovltechnicalreport}, and SEED1.5-VL~\citep{guo2025seed15vltechnicalreport}, \etc. These models adopt textual coordinate representations with standard next-token prediction, providing a direct comparison to our parallel box decoding paradigm. \textbf{(3) VLM-based detection and grounding specialists}, including Rex-Omni~\citep{jiang2025rexomni}, which is the most related work to ours targeting unified detection and grounding in a VLM framework. For GUI grounding, we also include several domain-specific expert models~\citep{liu_infigui-r1_2025, xie_scaling_2025, liu2025scalecua,yang_gta1_2025, ye_mobile-agent-v3_2025, zhou_mai-ui_2025, team_ui-venus-15_2026}.

\vspace{+1mm}
\noindent\textbf{Evaluation Setup.} Following the evaluation framework established in Rex-Omni~\citep{jiang2025rexomni}, we conduct a comprehensive assessment across multiple visual perception tasks.
Object Detection is evaluated on COCO for common objects, LVIS~\citep{gupta2019lvis} for long-tailed distributions, and VisDrone \citep{du2019visdrone} and Dense200~\citep{jiang2025rexomni} for dense and tiny object scenarios. Language-aware Grounding tasks include Referring Expression Comprehension (REC) on RefCOCOg~\citep{kazemzadeh-etal-2014-referitgame} and HumanRef~\citep{jiang2025referring}. Interactive tasks are evaluated through GUI Grounding on ScreenSpot-Pro~\citep{li2025screenspot}. Additionally, Layout Grounding on DocLayNet~\citep{pfitzmann2022doclaynet} and M6Doc~\citep{cheng2023m6doc}, along with OCR (text detection and recognition) on TotalText~\citep{ch2017total}, are reported together under scene text and document understanding tasks. 

The metric for each task is summarized as follows. \textit{(1) Box-based outputs}: For detection, layout, and OCR tasks, a prediction is considered correct (\ie, a true positive) if its Intersection over Union (IoU) with the ground truth exceeds a certain threshold. The F1-score is reported at $IoU = 0.5$, $IoU = 0.95$, and as a mean over thresholds ($mIoU$). \textit{(2) Point-based outputs}: For pointing tasks, a prediction is considered correct if the predicted point falls within the ground-truth segmentation mask or bounding box. We similarly report the F1-score for these point-based outputs based on this correctness criterion.

\begin{table*}[t]
    \centering
    \vspace{+3mm}
    \caption{Results on dense object detection benchmark Dense200 and VisDrone. }
    \resizebox{1.0\textwidth}{!}{
        \setlength{\tabcolsep}{10pt}
        \begin{tabular}{l|c|ccc|ccc}
            \toprule
            \multirow{3}{*}{\textbf{Method}} &
            \multirow{3}{*}{\begin{tabular}[c]{@{}c@{}}\textbf{Score} \\ \textbf{Thresh.}\end{tabular}} &
            \multicolumn{3}{c|}{\textbf{Dense200}} &
            \multicolumn{3}{c}{\textbf{VisDrone}} \\
            \cline{3-8}
             & &
             \begin{tabular}[c]{@{}c@{}}F1@IoU\\ 0.5\end{tabular} &
             \begin{tabular}[c]{@{}c@{}}F1@IoU\\ 0.95\end{tabular} &
             \begin{tabular}[c]{@{}c@{}}F1@IoU\\ Mean\end{tabular} &
             \begin{tabular}[c]{@{}c@{}}F1@IoU \\ 0.5\end{tabular} &
             \begin{tabular}[c]{@{}c@{}}F1@IoU\\ 0.95\end{tabular} &
             \begin{tabular}[c]{@{}c@{}}F1@IoU\\ Mean\end{tabular} \\
            \midrule

            \multicolumn{8}{c}{{Open-set Specialized Detectors}} \\
            \midrule
            Grounding DINO-Swin-T~\citep{liu2023grounding} & 0.25 & 36.9 & \textbf{19.7} & 33.1 & 55.2 & \textbf{3.9} & \underline{38.5} \\
            \midrule

            \multicolumn{8}{c}{{Vision-Language Models}} \\
            \midrule
            DeepSeek-VL2-Small~\citep{wu2024deepseekvl2mixtureofexpertsvisionlanguagemodels} & - & 16.0 & 3.9 & 12.7 & 35.8 & 1.7 & 23.3 \\
            OVIS2.5-2B~\citep{lu2025ovis25technicalreport}         & - & 17.9 & 0.0 & 6.7 & 21.0 & 0.1 & 9.2 \\
            MiMo-VL-7B~\citep{coreteam2025mimovltechnicalreport}         & - & 29.7 & 0.4 & 15.9 & 27.7 & 0.3 & 14.3 \\
            Qwen3-VL-4B~\citep{bai2025qwen3vltechnicalreport}        & - & 17.5 & 2.4 & 12.5 & 42.3 & 1.4 & 26.0 \\
            Qwen3-VL-8B~\citep{bai2025qwen3vltechnicalreport}        & - & 13.5 & 1.7 & 9.6  & 42.8 & 1.4 & 25.8 \\
            Cosmos-Reason2-8B~\citep{cosmosreason2}        & - & 25.1 & 1.1 & 15.1 & 40.2 & 1.3 & 22.3 \\
            SEED1.5-VL~\citep{guo2025seed15vltechnicalreport}         & - & \underline{76.9} & 5.3 & 53.2 & 55.9 & 0.6 & 27.4 \\
            Rex-Omni-SFT-3B~\citep{jiang2025rexomni}       & - & 60.2 & 10.6 & 46.4 & 55.6 & 1.9 & 32.4 \\
            Rex-Omni-3B~\citep{jiang2025rexomni}           & - & \textbf{78.4} & 10.3 & \underline{58.3} & \underline{61.6} & 1.5 & 35.8 \\
            \rowcolor{lightblue!10}
            LocateAnything-3B     & - & 74.0 & \underline{18.5} & \textbf{58.7} & \textbf{63.0} & \underline{3.2} & \textbf{39.9} \\
            \bottomrule
        \end{tabular}
    }
    \label{tab:dense_object_detection}
\end{table*}

\vspace{-2mm}
\subsection{Main Results}
In this section, we report the accuracy metrics and the throughput (measured in boxes per second, BPS on a single NVIDIA H100 GPU with a batch size of 1) of LocateAnything under the default \textit{Hybrid Mode}. The results of \textit{Fast} and \textit{Slow Mode} are provided in the \textbf{supplementary materials}.

\begin{table*}[t]
  \centering
  \caption{Results for the GUI Grounding task. The * denotes our reproduced results.}
  \resizebox{\textwidth}{!}{%
    \setlength{\tabcolsep}{6pt}
    \begin{tabular}{l|ccccccccccccc}
      \toprule
      \multirow{3}{*}{\textbf{Method}}
        & \multicolumn{13}{c}{\textbf{ScreenSpot-Pro}}
      \\ \cline{2-14}
        & \multicolumn{2}{c}{Dev.}
        & \multicolumn{2}{c}{Creative}
        & \multicolumn{2}{c}{CAD}
        & \multicolumn{2}{c}{Sci.}
        & \multicolumn{2}{c}{Office}
        & \multicolumn{2}{c}{OS}
        & \multirow{2}{*}{Avg}
      \\ \cline{2-13}
        & Text & Icon
        & Text & Icon
        & Text & Icon
        & Text & Icon
        & Text & Icon
        & Text & Icon
        &
      \\ \midrule
      InfiGUI-R1-3B~\citep{liu_infigui-r1_2025}
        & 51.3 & 12.4 & 44.9 & 7.0  & 33.0 & 14.1 & 58.3 & 20.0 & 65.5 & 28.3 & 43.9 & 12.4 & 35.7 \\
      JEDI-3B~\citep{xie_scaling_2025}
        & 61.0 & 13.8 & 53.5 & 8.4  & 27.4 & 9.4  & 54.2 & 18.2 & 64.4 & 32.1 & 38.3 & 9.0  & 36.1 \\
      Rex-Omni-3B~\citep{jiang2025rexomni}
        & 61.7 & 9.7  & 52.5 & 12.6 & 22.3 & 9.4  & 59.0 & 26.4 & 63.3 & 28.3 & 24.1 & 15.7 & 36.8 \\
      ScaleCUA-3B~\citep{liu2025scalecua}
        & 57.8 & 18.6 & 38.8 & \underline{42.9} & 16.8 & 32.0 & 54.3 & 28.1 & 47.9 & \underline{64.6} & 35.5 & \textbf{52.0} & 40.8 \\
      GTA1-7B~\citep{yang_gta1_2025}
        & 62.6 & 18.2 & 53.3 & 17.2 & \textbf{66.9} & 20.7 & 76.4 & 31.8 & \underline{82.5} & 50.9 & 48.6 & 25.9 & 50.1 \\
      Qwen3-VL-30B-A3B*~\citep{bai2025qwen3vltechnicalreport}
        & 76.0 & 24.8 & \textbf{69.2} & 20.3 & 51.8 & 15.6 & 76.4 & 27.3 & 80.8 & 37.7 & \textbf{75.7} & 38.2 & 53.7 \\
      GUI-Owl-7B~\citep{ye_mobile-agent-v3_2025}
        & \underline{76.6} & 31.0 & 59.6 & 27.3 & \underline{64.5} & 21.9 & 79.1 & 37.3 & 77.4 & 39.6 & 59.8 & 33.7 & 54.9 \\
      MAI-UI-2B~\citep{zhou_mai-ui_2025}
        & \underline{76.6} & 32.4 & \textbf{69.2} & 21.7 & 61.4 & 23.4 & \underline{81.2} & 34.5 & \textbf{85.9} & 39.6 & 68.2 & 41.6 & 57.4 \\
      UI-Venus-1.5-2B~\citep{team_ui-venus-15_2026}
        & 70.1 & \underline{43.4} & 63.6 & 28.7 & 54.3 & \underline{32.8} & 76.4 & 38.2 & 81.9 & 47.2 & \underline{73.8} & \underline{51.7} & 57.7 \\
      GUI-Owl-32B~\citep{ye_mobile-agent-v3_2025}
        & \textbf{84.4} & 39.3 & \underline{65.2} & 18.2 & 62.4 & 28.1 & \textbf{82.6} & \underline{39.1} & 81.4 & 39.6 & 70.1 & 36.0 & \underline{58.0} \\
\rowcolor{lightblue!10}
      LocateAnything-3B
        & 70.8 & \textbf{50.3} & 60.1 & \textbf{46.9}
        & 57.9 & \textbf{40.6} & 69.4 & \textbf{58.2}
        & 77.2 & \textbf{69.8} & 65.4 & 43.8
        & \textbf{60.3}
      \\

      \bottomrule
    \end{tabular}%
  }
  \label{tab:gui_grounidng}
\end{table*}

\vspace{+1mm}
\noindent\textbf{High-Quality Multi-Object Detection.}
Our model exhibits robust generalization in both common and complex dense object detection scenarios. On general detection benchmarks reported in Tab.~\ref{tab:lvis_coco_f1_merged}, LocateAnything improves the mean F1 by +3.8\% on LVIS and +1.8\% on COCO compared to Rex-Omni, despite sharing an identical model size. Crucially, the model effectively learns the generalized spatial distribution, transferring its detection capabilities to unseen, heavily packed object types. This is evidenced by its performance on the dense detection benchmarks in Tab.~\ref{tab:dense_object_detection}, where it achieves 39.9 mean F1 on VisDrone, substantially outperforming Rex-Omni which scores 35.8. Similarly, it reaches a competitive 58.7 mean F1 on Dense200, demonstrating superior boundary delineation and instance separation in heavily overlapping environments.

\begin{table*}[t]
    \centering
    \vspace{+3mm}
    \caption{Performance comparison on document layout grounding and OCR tasks.}
    \resizebox{0.98\textwidth}{!}{
        \setlength{\tabcolsep}{6pt}
        \begin{tabular}{l|c|ccc|ccc|ccc}
\toprule
\multirow{3}{*}{\textbf{Method}} &
\multirow{3}{*}{\begin{tabular}[c]{@{}c@{}}\textbf{Score}\\ \textbf{ Thresh.}\end{tabular}} &
\multicolumn{3}{c|}{\textbf{DocLayNet}} &
\multicolumn{3}{c|}{\textbf{M6Doc}} &
\multicolumn{3}{c}{\textbf{TotalText}} \\
\cline{3-11}
& &
\begin{tabular}[c]{@{}c@{}}F1@IoU\\ 0.5\end{tabular} &
\begin{tabular}[c]{@{}c@{}}F1@IoU\\ 0.95\end{tabular} &
\begin{tabular}[c]{@{}c@{}}F1@IoU\\ Mean\end{tabular} &
\begin{tabular}[c]{@{}c@{}}F1@IoU\\ 0.5\end{tabular} &
\begin{tabular}[c]{@{}c@{}}F1@IoU\\ 0.95\end{tabular} &
\begin{tabular}[c]{@{}c@{}}F1@IoU\\ Mean\end{tabular} &
\begin{tabular}[c]{@{}c@{}}F1@IoU\\ 0.5\end{tabular} &
\begin{tabular}[c]{@{}c@{}}F1@IoU\\ 0.95\end{tabular} &
\begin{tabular}[c]{@{}c@{}}F1@IoU\\ Mean\end{tabular} \\
\midrule

\multicolumn{11}{c}{{Specialized Detectors}} \\
\midrule
DocLayout-YOLO~\citep{zhao2024doclayout} & 0.3
& \textbf{91.2} & \textbf{52.1} & \textbf{81.1}
& - & - & -
& - & - & - \\
PaddleOCRv5~\citep{cui2025paddleocr} & -
& - & - & -
& - & - & -
& 40.2 & 0.7 & 25.7 \\
\midrule

\multicolumn{11}{c}{{Vision-Language Models}} \\
\midrule
SEED1.5-VL~\citep{guo2025seed15vltechnicalreport} & -
& 54.9 & 4.3 & 28.7
& 48.0 & 3.4 & 28.0
& 35.0 & 0.3 & 19.5 \\
Qwen3-VL-4B~\citep{bai2025qwen3vltechnicalreport} & -
& 60.8 & 8.2 & 37.2
& 30.6 & 4.9 & 19.0
& 55.4 & 3.6 & 36.1 \\
Qwen3-VL-8B~\citep{bai2025qwen3vltechnicalreport} & -
& 54.7 & 6.7 & 34.1
& 37.2 & 4.9 & 22.7
& \textbf{59.4} & 2.7 & 37.3 \\
Rex-Omni-3B~\citep{jiang2025rexomni} & -
& 89.5 & 28.4 & 70.7
& \underline{76.3} & \underline{18.7} & \underline{55.6}
& 56.6 & \underline{3.9} & \underline{40.6} \\
\rowcolor{lightblue!10}
LocateAnything-3B & -
& \underline{91.1} & \underline{35.8} & \underline{76.8}
& \textbf{90.6} & \textbf{25.8} & \textbf{70.1}
& \underline{58.9} & \textbf{5.1} & \textbf{43.3} \\
\bottomrule
        \end{tabular}
    }
    \label{tab:layout_ocr_merged}
\end{table*}

\vspace{+1mm}
\noindent\textbf{Precise Open-World Localization Ability.}
LocateAnything demonstrates exceptional fine-grained localization capabilities across diverse open-world benchmarks, including user interface grounding, document layout parsing, and referring expression comprehension. As shown in Tab.~\ref{tab:gui_grounidng}, on the ScreenSpot-Pro~\citep{li2025screenspot}, it achieves a SOTA mean F1 of 60.3, surpassing generalist VLMs like Qwen3-VL-30B-A3B and specialized models tailored for UI tasks such as GUI-Owl-32B. Furthermore, in document understanding tasks detailed in Tab.~\ref{tab:layout_ocr_merged}, LocateAnything establishes a new standard by reaching 76.8 and 70.1 mean F1 on DocLayNet and M6Doc, respectively, outperforming Rex-Omni by substantial margins. This precise spatial reasoning extends to complex referring tasks, as shown in Tab.~\ref{tab:referring_detection}, where the model seamlessly aligns nuanced human intents with visual regions, achieving 78.7 mean F1 on the HumanRef benchmark and remaining highly competitive on RefCOCOg against top-tier models.

\vspace{+1mm}
\noindent\textbf{Superior Decoding Speed.}
A key advantage of our model is its drastically reduced decoding steps. As shown in Tab.~\ref{tab:lvis_coco_f1_merged}, our model achieves 12.7 BPS under the default hybrid mode, over 10$\times$ faster than textual-based Qwen3-VL (1.1 BPS) and 2.5$\times$ faster than quantized-based Rex-Omni (5.0 BPS).

\begin{table*}[t]
    \caption{Evaluation results on referring expression comprehension benchmarks.}
    \centering
    \resizebox{1.0\textwidth}{!}{
        \setlength{\tabcolsep}{4pt}
        \begin{tabular}{l|c|ccc|ccc|ccc}
\toprule
\multirow{3}{*}{\textbf{Method}} & 
\multirow{3}{*}{\begin{tabular}[c]{@{}c@{}}\textbf{Score} \\ \textbf{Thresh.}\end{tabular}} 
& \multicolumn{3}{c|}{\textbf{HumanRef}} 
& \multicolumn{3}{c|}{\textbf{RefCOCOg val}} 
& \multicolumn{3}{c}{\textbf{RefCOCOg test}} \\
\cline{3-11}
 &  & \begin{tabular}[c]{@{}c@{}}F1@IoU\\ 0.5\end{tabular} & \begin{tabular}[c]{@{}c@{}}F1@IoU\\ 0.95\end{tabular} & \begin{tabular}[c]{@{}c@{}}F1@IoU\\ Mean\end{tabular} 
 & \begin{tabular}[c]{@{}c@{}}F1@IoU\\ 0.5\end{tabular} & \begin{tabular}[c]{@{}c@{}}F1@IoU\\ 0.95\end{tabular} & \begin{tabular}[c]{@{}c@{}}F1@IoU\\ Mean\end{tabular} 
 & \begin{tabular}[c]{@{}c@{}}F1@IoU\\ 0.5\end{tabular} & \begin{tabular}[c]{@{}c@{}}F1@IoU\\ 0.95\end{tabular} & \begin{tabular}[c]{@{}c@{}}F1@IoU\\ Mean\end{tabular} \\
\midrule

\multicolumn{11}{c}{{Open-set Specialized Detector}} \\
\midrule
Grounding DINO-Swin-T~\citep{liu2023grounding} & 0.25 
& 28.0 & 16.5 & 25.2 
& 52.9 & 20.9 & 45.9 
& 53.8 & 22.9 & 46.8 \\
\midrule
\multicolumn{11}{c}{{Vision-Language Model}} \\
\midrule
DeepSeek-VL2-Tiny~\citep{wu2024deepseekvl2mixtureofexpertsvisionlanguagemodels} & - 
& 39.1 & 16.9 & 31.4 
& 67.4 & 16.1 & 50.5 
& 69.3 & 16.9 & 52.1 \\
OVIS2.5-2B~\citep{lu2025ovis25technicalreport} & - 
& 70.6 & 12.3 & 50.0 
& 87.4 & 29.3 & 73.4 
& 87.6 & 30.5 & 73.8 \\
MiMo-VL-7B~\citep{coreteam2025mimovltechnicalreport} & - 
& 77.6 & 26.4 & 63.4 
& 84.9 & 14.4 & 65.3 
& 84.6 & 14.9 & 65.5 \\
DeepSeek-VL2-Small~\citep{wu2024deepseekvl2mixtureofexpertsvisionlanguagemodels} & - 
& 72.0 & 46.5 & 64.7 
& \textbf{92.4} & \textbf{45.6} & \textbf{81.4} 
& \textbf{91.8} & \textbf{47.0} & \textbf{81.6} \\
Qwen3-VL-4B~\citep{bai2025qwen3vltechnicalreport} & - 
& 77.7 & 54.9 & 71.1
& 88.0 & 34.0 & 74.7
& 87.6 & 33.9 & 74.6 \\
Qwen3-VL-8B~\citep{bai2025qwen3vltechnicalreport} & -
& 78.6 & 55.7 & 72.0 
& \underline{88.6} & 33.4 & 74.9 
& 88.6 & 33.8 & 75.2 \\
SEED1.5-VL~\citep{guo2025seed15vltechnicalreport} & - 
& \textbf{88.2} & 60.0 & \textbf{81.6} 
& 84.7 & 30.9 & 71.9 
& 85.2 & 32.1 & 73.2 \\
Rex-Omni-3B~\citep{jiang2025rexomni} & - 
& \underline{85.4} & \underline{65.4} & \underline{79.9} 
& 86.6 & 35.3 & 73.6 
& 86.8 & 36.6 & 74.3 \\
\rowcolor{lightblue!10}
LocateAnything-3B & - 
& 82.9 & \textbf{68.8} & 78.7 
& \underline{88.6} & \underline{41.5} & \underline{76.7} 
& \underline{88.8} & \underline{43.4} & \underline{77.6} \\
\bottomrule
\end{tabular}
    }
    \label{tab:referring_detection}
\end{table*}

\vspace{-2mm}
\subsection{Ablation Study}
\begin{figure}[t]
\centering
\includegraphics[width=\linewidth]{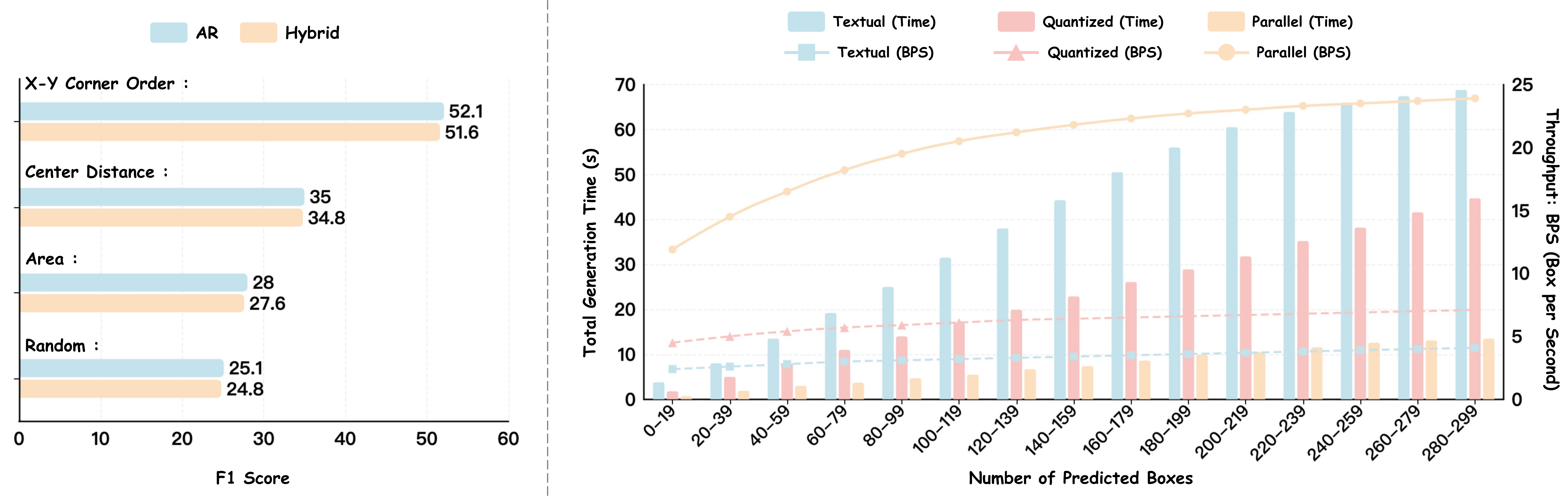}
\caption{\textbf{Ablation Study on Box Ordering and Decoding Speed.} \textbf{Left:} Effect of different box sorting strategies on the F1-score. \textbf{Right:} Comparison of Generation Time (bars) and Throughput (lines) across varying numbers of predicted boxes for Textual, Quantized, and Parallel box decoding.}
\label{fig:chart}
\vspace{+2mm}
\end{figure}

\vspace{-1mm}
We conduct ablation studies on the COCO dataset to validate our core designs. The results are shown in Tab.~\ref{tab:combined_ablation} and Fig.~\ref{fig:chart}.

\noindent\textbf{Coordinate Representation.} As Tab.~\ref{tab:combined_ablation}(a) shows, under the NTP paradigm, Textual and Quantized representations yield sub-optimal performance (49.1 and 50.1 mean F1, respectively) due to forced token-by-token generation. Our PBD (Slow Mode) achieves the highest F1-score of 52.1, proving that a box-aligned formulation provides stronger supervision for spatial reasoning than 1D serialization, without sacrificing throughput.

\noindent\textbf{MTP Formulation.} Tab.~\ref{tab:combined_ablation}(b) compares our box-aligned MTP against existing structure-agnostic MTP formulations. Methods like SDLM and Block Diffusion force the model to learn spurious, unaligned cross-boundary patterns, suffering from lower accuracy and limited acceleration (\eg, SDLM-B6 achieves 46.1 F1-score at 5.5 BPS). Furthermore, structure-agnostic methods (\eg, SDLM-B4, B6, B8) exhibit a strict speed-accuracy trade-off, where increasing the block size yields only marginal throughput gains while consistently degrading the F1-score. In contrast, our PBD strictly aligns MTP blocks with structured bounding box units, dramatically outpacing existing methods in throughput (16.9 BPS) while improving the mean F1 to 49.6.

\noindent\textbf{Decoding Mode.} Tab.~\ref{tab:combined_ablation}(c) ablates the impact of our dual-formulation training ($\mathcal{L}_{\mathrm{ntp}}$ and $\mathcal{L}_{\mathrm{blk}}$). Training with isolated losses limits the model's potential; joint training successfully pushes the Slow Mode upper bound from 50.1 to 52.1 F1-score. During inference, Fast Mode (MTP) maximizes throughput (16.9 BPS) but induces accuracy drops in complex scenes. Hybrid Mode seamlessly resolves this trade-off, preserving most speed gains (13.2 BPS) while achieving robust, high-precision localization (51.6 F1-score).

\noindent\textbf{Box Output Order.} We investigate four spatial sorting strategies in Fig.~\ref{fig:chart} (left): \textit{X-Y Corner Order} (sorting by the x-coordinate of the left-top corner, then by the y-coordinate), \textit{Center Distance} (the distance of the bounding box center point to the origin), \textit{Area} (sorted from largest to smallest), and \textit{Random} (shuffled randomly). Results show X-Y Corner Order yields the highest F1-score. We take this setting as  default in dataset construction.

\begin{table*}[t]
\centering
\vspace{+3mm}
\caption{\small Ablation Studies on the COCO dataset. We decouple the analysis into three aspects: (a) coordinate representation, (b) block-based MTP Formulation, and (c) effectiveness of our on-demand decoding modes and loss design. Throughput is measured in boxes per second. For brevity, we report the Average metric across IoU thresholds for Recall (R), Precision (P), and F1 Score. ``B'' indicates block size in MTP.}
\label{tab:combined_ablation}

\resizebox{\textwidth}{!}{%
\begin{tabular}[c]{l cccc}
\multicolumn{5}{c}{\textbf{(a) Coordinate Representations}} \\
\toprule
Method & Throughput & R & P & F1 \\
\midrule
Textual & 1.3 & 45.7 & 52.3 & 49.1 \\
Quantized & 3.9 & 48.2 & 52.2 & 50.1 \\
\midrule
PBD (Slow) & 3.9 & \textbf{49.4} & \textbf{55.2} & \textbf{52.1} \\
PBD (Fast) & \textbf{16.9} & 45.6 & 54.6 & 49.6 \\
PBD (Hybrid) & \underline{13.2} & \underline{48.7} & \underline{54.8} & \underline{51.6} \\
\bottomrule
\end{tabular}%
\hspace{4mm}%
\begin{tabular}[c]{l cccc}
\multicolumn{5}{c}{\textbf{(b) MTP Formulations}} \\
\toprule
Method & Throughput & R & P & F1 \\
\midrule
SDLM-B4~\citep{liu2025sequential} & 5.2 & \textbf{45.4} & \underline{48.1} & \underline{46.5} \\
SDLM-B6~\citep{liu2025sequential} & 5.5 & 45.1 & 47.5 & 46.1 \\
SDLM-B8~\citep{liu2025sequential} & \underline{6.7} & 44.7 & 47.2 & 45.8 \\
\midrule
Block Diff-B6~\citep{arriola2025block} & 4.7 & 45.1 & 44.3 & 44.8 \\
\midrule
PBD (Fast) & \textbf{16.9} & \underline{45.2} & \textbf{54.6} & \textbf{49.6} \\
\bottomrule
\end{tabular}%
\hspace{4mm}%
\begin{tabular}[c]{cc l cccc}
\multicolumn{7}{c}{\textbf{(c) Decoding Modes \& Losses}} \\
\toprule
$\mathcal{L}_{\mathrm{ntp}}$ & $\mathcal{L}_{\mathrm{blk}}$ & Mode & Throughput & R & P & F1 \\
\midrule
\checkmark & & Slow & 3.9 & 48.2 & 52.2 & 50.1 \\
& \checkmark & Fast & \underline{16.7} & 45.6 & 49.0 & 47.2 \\
\midrule
    \checkmark & \checkmark & Slow & 3.9 & \textbf{49.4} & \textbf{55.2} & \textbf{52.1} \\
\checkmark & \checkmark & Fast & \textbf{16.9} & 45.6 & 54.6 & 49.6 \\
\checkmark & \checkmark & Hybrid & 13.2 & \underline{48.7} & \underline{54.8} & \underline{51.6} \\
\bottomrule
\end{tabular}%
}
\end{table*}

\noindent\textbf{Throughput.} 
We compare generation time and throughput with NTP methods in Fig.~\ref{fig:chart} (right).
As target boxes increase from 20 to 300, NTP methods suffer from a severe latency bottleneck. In contrast, the Parallel method exhibits little increase in generation time, increasing throughput from 12 BPS to $\sim$25 BPS in dense scenes. These findings confirm that PBD effectively breaks the decoding bottleneck, achieving a $2\times$ to $6\times$ speedup.

\subsection{Qualitative Results}
\begin{figure}[t!]
    \centering
    \includegraphics[width=\linewidth]{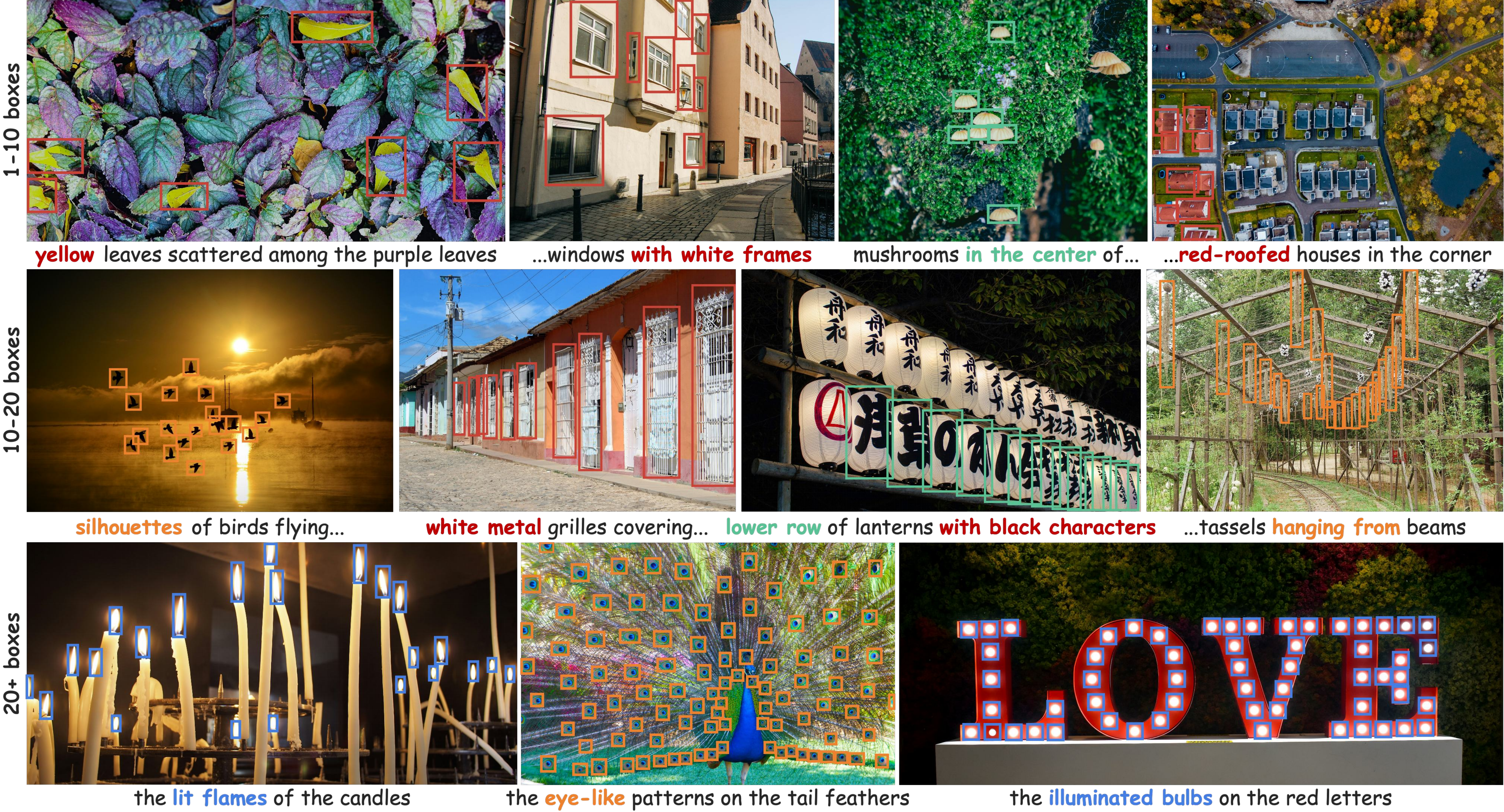} 
    \caption{\textbf{Qualitative results}. Each row shows test cases with varying numbers of target objects and diverse box scales. Different colors indicate different query categories, including \textcolor{MyRed}{attribute}, \textcolor{MyBlue}{part}, \textcolor{MyOrange}{reasoning}, and \textcolor{MyGreen}{spatial} queries. Our model consistently localizes targets across diverse scene domains, arbitrary image resolutions, free-form textual queries, and an arbitrary number of objects, demonstrating strong robustness.
}
    \label{fig:vis}
\end{figure}

Fig.~\ref{fig:vis} visualizes representative grounding results of our model. Visual comparisons with other methods are provided in the \textbf{supplementary materials}. We observe three consistent behaviors.
\textbf{(i) Compositional grounding:} our model handles attribute/part/spatial/reasoning-style queries well with consistent spatial alignment, supported by the diversity and coverage of our training data.
\textbf{(ii) Robustness to large instance counts:} as targets grow from sparse to crowded settings, the predicted boxes remain structured and accurate, reflecting the precision of our box-level decoding. This robustness is further strengthened by our Stage-2 training that emphasizes many-object images, improving dense localization in practice. Moreover, our Hybrid Mode maintains most of the parallel decoding speed while improving output stability in multi-instance generation.
\textbf{(iii) Reliable localization in clutter:} boxes stay compact and well-separated under occlusion, repetitive textures, and grid-like dense layouts. Our hybrid inference mode further stabilizes these hard cases by detecting unreliable parallel blocks and falling back to NTP re-decoding when needed.

\section{Conclusion}\label{sec:conclusion}
We presented LocateAnything, a unified framework that reformulates visual grounding and detection in VLMs via \emph{Parallel Box Decoding}. By elevating geometric elements to atomic units rather than 1D streams, LocateAnything aligned the training supervision with the inherently coupled nature of spatial coordinates. With massive 138M text-image training queries and a flexible on-demand inference mechanism, LocateAnything not only delivered SOTA accuracy across diverse tasks, but also achieved up to a $2.5\times$ speedup over competitive methods. Our method provided a practical and scalable route for real-time visual perception, opening the door to deploying general-purpose VLMs in latency-sensitive embodied robotics and interactive agents.

\noindent\textbf{Limitation.} Currently, our model is primarily trained with supervised fine-tuning. Reinforcement learning is an important next step to further optimize the block-level decoding policy, reduce fallback frequency, and encourage effective exploration in hard dense/long-tail cases, which could improve both robustness and worst-case decoding speed. We leave it for future work.

\noindent\textbf{Acknowledgement.} The authors would like to thank the valuable discussions and input from Qing Jiang, Amala Sanjay Deshmukh, Karan Sapra, Mingjie Liu, Yi Dong, Pavlo Molchanov, Yonggan Fu, Collin McCarthy, Mike Ranzinger, Greg Heinrich, Wonmin Byeon, Yexuan Li, Chi-Pin Huang, Fu-En Yang, Frank Wang, Jin Huang, Le An, Jaehun Jung, Shaokun Zhang, Hao Zhang, Johan Bjoerck, Jim Fan, Patrick Langechuan Liu, Sifei Liu, Xiaolong Li, Paris Zhang, Yilin Zhao, Subhashree Radhakrishnan, Shiyi Lan, Jose Alvarez, Sanja Fidler, Yan Wang, Xiaodong Yang, Yin Cui, Tsung-Yi Lin, Padmavathy Subramanian and more. We would also like to thank the NVIDIA infra, legal and data teams, including Xinyou Ma, Katherine Cheung, Timo Roman, and Yao Xu for their prompt and helpful support. Finally, the authors would like to additionally acknowledge the following teams, including Nemotron-Diffusion, Nemotron VLM, Cosmos, GR00T, Alpamayo, Gigas and Metropolis, for the engagement and downstream applications.

\clearpage
\appendix

\section{Training and Inference Configurations}

\subsection{Training Details}

In this section, we provide extended details regarding our data mixture strategy, the multiphase training pipeline of our base VLM and the LocateAnything model. We also elaborate on two key system-level techniques that are critical for efficient training under our dual-formulation design: \textit{Stream Packing} for maximizing GPU utilization, and \textit{MagiAttention}~\citep{magiattention2025} for natively supporting the heterogeneous attention masks required by our NTP+MTP joint training.

To provide a comprehensive overview of our entire training pipeline, Tab.~\ref{tab:details_sft} summarizes the detailed optimization hyperparameters and configurations across all four progressive stages of \textbf{LocateAnything}.

\begin{table}[!tbp]
    \vspace{4mm}
    \caption{Detailed configuration for each training stage of \textbf{LocateAnything}.}
    \centering
    \setlength{\tabcolsep}{12pt}
    \renewcommand{\arraystretch}{1.2}
    \resizebox{0.98\textwidth}{!}{%
    \begin{tabular}{@{}l|c|c|c|c}
        \toprule
        \textbf{Stages} & \textbf{Stage 1} & \textbf{Stage 2} & \textbf{Stage 3} & \textbf{Stage 4} \\ 
        \midrule
        \textbf{Objective} & \multicolumn{2}{c|}{\makecell[c]{\textbf{World Knowledge} \\ \textbf{Injection}}} & \multicolumn{2}{c}{\makecell[c]{\textbf{Detection \& Grounding} \\ \textbf{Enhancement}}} \\
        \midrule
        \textbf{Dataset} & \makecell[c]{Caption} & \makecell[c]{General VQA} & \makecell[c]{Detection \& Grounding} & \makecell[c]{20\% Previous  + Dense} \\
        \midrule
        \textbf{Learning Rate} & 2 $\times 10^{-4}$ & 4 $\times 10^{-5}$ & 4 $\times 10^{-5}$ & 1 $\times 10^{-5}$ \\
        \textbf{Optimizer} & AdamW & AdamW & AdamW & AdamW \\
        \textbf{Weight Decay} & 0.01 & 0.01 & 0.01 & 0.01 \\
        \textbf{LR Schedule} & Cosine & Cosine & Cosine & Cosine \\
        \textbf{Max Sequence Length} & 32768 & 32768 & 25600 & 25600 \\
        \textbf{Trainable Components} & MLP & All & All & All \\
        \midrule
        \textbf{Number of GPUs} & 64 & 256 & 256 & 256 \\
        \textbf{Training Steps} & 2000 & 20000 & 25000 & 5000 \\
        \bottomrule
    \end{tabular}
    }
    \label{tab:details_sft}
\end{table}

\subsubsection{Base VLM Training (World Knowledge Injection)}
To establish a robust foundational understanding of world knowledge before introducing specialized detection and grounding tasks, we first pretrain our base VLM. This initial alignment phase strictly excludes any detection or bounding-box grounding data and is divided into the first two progressive stages. (1) \textbf{Stage 1 (Visual Concept Initialization):} In this stage, the model is trained exclusively on caption-related datasets, detailed in the ``Captioning \& Knowledge'' category of Tab.~\ref{tab:data_sft}. This enables the native any-resolution visual encoder to align fundamental visual features with textual descriptions effectively. (2) \textbf{Stage 2 (Comprehensive Multimodal Learning):} Building upon the basic captioning capability, we expand the training corpus to encompass all datasets listed in Tab.~\ref{tab:data_sft}. This comprehensive mixture spans a wide spectrum of domains, including Mathematics \& Code, Science, Chart \& Table reasoning, extensive OCR tasks (Naive OCR and OCR QA), General VQA, Text-only instruction tuning, and basic Counting. Fully integrating these diverse datasets ensures the base model develops strong reasoning and comprehensive multimodal capabilities.

\begin{table*}[!tbp]
\caption{\textbf{Datasets used for the initial world-knowledge alignment.} We pretrain the base VLM on this diverse mixture of datasets across various domains to ensure broad coverage of general knowledge. Specifically, Stage-1 incorporates only the caption-related datasets shown here. In Stage-2, all datasets listed in this table are fully integrated into the training process to build robust, comprehensive multimodal capabilities.}
\renewcommand{\arraystretch}{1.2}
    \setlength\tabcolsep{6pt}
    \resizebox{\textwidth}{!}{
    \begin{tabular}{m{1.6cm}|>{\small}m{15.4cm}}
\footnotesize Category & \small Dataset  \\
\hline

\rowcolor{gray!15}\footnotesize 
Captioning \& Knowledge & ShareGPT4o~\cite{opengvlab_sharegpt4o_dataset}, KVQA~\cite{shah2019kvqa}, Movie-Posters~\cite{skvarre_movie_posters_100k}, Google-Landmark~\cite{weyand2020googlelandmark}, WikiArt~\cite{wikiart_dataset}, Weather-QA~\cite{ma2024weatherqa}, Coco-Colors~\cite{mscoco-controlnet-canny-less-colors}, music-sheet~\cite{sheet_music_clean}, SPARK~\cite{yu2024spark}, Image-Textualization~\cite{pi2024image_textualization}, SAM-Caption~\cite{pixart_alpha_sam_llava_captions10m}, Tmdb-Celeb-10k~\cite{ashraq_tmdb_celeb_10k}, CC3M~\cite{sharma2018conceptual}, pixmo-cap~\cite{deitke2025molmo}, Multi-UI~\cite{liu2024harnessingwebpageuistextrich}, RICO~\cite{deka2017rico} \\

\footnotesize 
Mathematics \& Code & GeoQA+~\cite{cao2022geoqa_plus}, MathQA~\cite{yu2023mathqa}, CLEVR-Math/Super~\cite{lindstrom2022clevrmath, li2023superclevr}, Geometry3K~\cite{lu2021intergps}, MAVIS-math-rule-geo~\cite{zhang2024mavis}, MAVIS-math-metagen~\cite{zhang2024mavis}, InterGPS~\cite{lu2021intergps}, Raven~\cite{zhang2019raven}, GEOS~\cite{seo2015geos}, UniGeo~\cite{chen2022unigeo}, Design2Code~\cite{si2025design2code}, OpenMathInstruct~\cite{toshniwal2024openmathinstruct}\\

\rowcolor{gray!15}\footnotesize 
Science &AI2D~\cite{kembhavi2016ai2d}, ScienceQA~\cite{lu2022scienceqa}, TQA~\cite{kembhavi2017tqa}, PathVQA~\cite{he2020pathvqa}, SciQA~\cite{auer2023sciqa}, Textbooks-QA, VQA-RAD~\cite{lau2018vqarad}, VisualWebInstruct~\cite{tiger_lab_visualwebinstruct}, PMC-VQA~\cite{zhang2023pmcvqa}

\\\footnotesize 
Chart \& Table &ChartQA~\cite{masry2022chartqa}, MMC-Inst~\cite{liu2023mmcinst}, DVQA~\cite{kafle2018dvqa}, PlotQA~\cite{methani2020plotqa}, LRV-Instruction~\cite{liu2023lrv-instruction}, TabMWP~\cite{lu2022tablemwp}, UniChart~\cite{masry2023unichart}, Vistext~\cite{tang2023vistext}, TAT-DQA~\cite{zhu2022tatdqa}, VQAonBD~\cite{VQAonDB}, FigureQA~\cite{kahou2017figureqa}, Chart2Text~\cite{kantharaj2022chart2text}, RobuT-\{Wikisql, SQA, WTQ\}~\cite{zhao2023robut}, MultiHiertt~\cite{zhao2022multihiertt}, MMTab~\cite{zheng2024multimodal} \\
\rowcolor{gray!15}\footnotesize

Naive OCR & SynthDoG~\cite{kim2022synthdog}, MTWI~\cite{he2018icpr2018_MTWI}, LVST~\cite{sun2019lsvt}, SROIE~\cite{huang2019icdar_sroie}, FUNSD~\cite{jaume2019funsd}, Latex-Formula~\cite{oleehyo_latex_formulas}, IAM~\cite{marti2002iam}, Handwriting-Latex~\cite{aida}, ArT~\cite{chng2019art}, CTW~\cite{yuan2019ctw}, ReCTs~\cite{zhang2019rects}, COCO-Text~\cite{veit2016cocotext}, SVRD~\cite{yu2023icdar_svrd}, Hiertext~\cite{long2023icdar_hiertext}, RoadText~\cite{tom2023icdar_roadtext}, MapText~\cite{li2024icdar_maptext}, CAPTCHA~\cite{captcha}, Est-VQA~\cite{wang2020estvqa}, HME-100K~\cite{tal}, TAL-OCR-ENG~\cite{tal}, TAL-HW-MATH~\cite{tal}, IMGUR5K~\cite{krishnan2023textstylebrush_Imgur5K}, ORAND-CAR~\cite{diem2014icfhr_RAND_CAR}, Invoices-and-Receipts-OCR~\cite{mychen76_invoices_receipts_ocr_v1},  Chrome-Writting~\cite{mouchere2016icfhr2016_chrome_writing}, IIIT5k~\cite{mishra2012scene_iiit5k}, K12-Printing~\cite{tal}, Memotion~\cite{ramamoorthy2022memotion}, Arxiv2Markdown, Handwritten-Mathematical-Expression~\cite{Azu}, WordArt~\cite{xie2022toward_wordart}, 
RenderedText~\cite{wendlerc_renderedtext}, Handwriting-Forms~\cite{ift_handwriting_forms}\\

\footnotesize 
OCR QA &  DocVQA~\cite{clark2017docqa}, InfoVQA~\cite{mathew2022infographicvqa}, TextVQA~\cite{singh2019textvqa}, ArxivQA~\cite{li2024multimodal_arxivQA},
ScreencQA~\cite{hsiao2022screenqa}, DocReason~\cite{mplug_docreason25k}, Ureader~\cite{ye2023ureader}, FinanceQA~\cite{Sujet-Finance-QA-Vision-100k}, DocMatrix~\cite{laurenccon2024building_docmatrix}, A-OKVQA~\cite{schwenk2022aokvqa}, Diagram-Image-To-Text~\cite{kamizuru00_diagram_image_to_text}, MapQA~\cite{chang2022mapqa}, OCRVQA~\cite{mishra2019ocrvqa}, ST-VQA~\cite{biten2019stvqa}, SlideVQA~\cite{tanaka2023slidevqa}, PDF-VQA~\cite{ding2023PDFvqa}, SQuAD-VQA, VQA-CD~\cite{mahamoud2024chic_vqa_cd}, Block-Diagram~\cite{shreyanshu09_block_diagram}, MTVQA~\cite{tang2024mtvqa}, ColPali~\cite{faysse2024colpali}, BenthamQA~\cite{mathew2021asking_benthamqa}, VSR~\cite{zhang2021vsr}, pixmo-docs~\cite{deitke2025molmo}\\

\rowcolor{gray!15}\footnotesize 
General VQA & LLaVA-150K~\cite{liu2023llava}, LVIS-Instruct4V~\cite{wang2023lvisinstruct4v}, ALLaVA~\cite{chen2024allava},  Laion-GPT4V~\cite{laion_gpt4v_dataset}, LLAVAR~\cite{zhang2023llavar}, SketchyVQA~\cite{tu2023many}, VizWiz~\cite{gurari2018vizwiz}, IDK~\cite{cha2024visually}, AlfworldGPT, LNQA~\cite{pont2020connecting_lnqa}, Face-Emotion~\cite{fastjob_visual_emotional_analysis}, SpatialSense~\cite{yang2019spatialsense}, Indoor-QA~\cite{keremberke_indoor_scene_classification}, Places365~\cite{zhou2017places365}, MMinstruct~\cite{liu2024mminstruct}, DriveLM~\cite{sima2023drivelm}, YesBut~\cite{nandy2024yesbut}, WildVision~\cite{lu2024wildvision}, LLaVA-Critic-113k~\cite{xiong2024llava_critic}, PhyCritic~\cite{xiong2026phycritic}, RLAIF-V~\cite{yu2024rlaif_v}, VQAv2~\cite{goyal2017vqav2}, MMRA~\cite{wu2024mmra}, KONIQ~\cite{hosu2020koniq}, MMDU~\cite{liu2024mmdu}, Spot-The-Diff~\cite{jhamtani2018learning_spotthediff}, Hateful-Memes~\cite{kiela2020hateful_memes}, COCO-QA~\cite{ren2015exploring_cocoqa}, NLVR~\cite{suhr2017corpus_nlvr2}, Mimic-CGD~\cite{laurenccon2024matters}, Datikz~\cite{belouadi2023automatikz_datikz},
Chinese-Meme~\cite{emo_visual_data_chinese_meme}, IconQA~\cite{lu2021iconqa}, Websight~\cite{laurenccon2024unlocking_websight}, OmniAlign~\cite{zhao2025omnialign}, pixmo-cap-qa~\cite{deitke2025molmo}, pixmo-ask-model-anything~\cite{deitke2025molmo}, Cauldron~\cite{laurenccon2024matters} \\

\footnotesize 
Text-only & Orca~\cite{lian2023openorca}, Orca-Math~\cite{mitra2024orca}, OpenCodeInterpreter~\cite{zheng2024opencodeinterpreter}
MathInstruct~\cite{yue2023mammoth_mathinstruct}, WizardLM~\cite{xu2023wizardlm}, TheoremQA~\cite{chen2023theoremqa}, OpenHermes2.5~\cite{OpenHermes2_5}, NuminaMath-CoT~\cite{numina_math_datasets}, Python-Code-25k~\cite{flytech_python_codes_25k}, Infinity-Instruct~\cite{baai_infinity_instruct},
Python-Code-Instructions-18k-Alpaca~\cite{iamtarun_python_code_instructions_18k_alpaca}, Ruozhiba~\cite{looksjuicy_ruozhiba}, InfinityMATH~\cite{zhang2024infinitymath}, StepDPO~\cite{lai2024stepDPO}, TableLLM~\cite{zhang2024tablellm}, UltraInteract-sft~\cite{yuan2024advancing_ultrainteract}\\

\rowcolor{gray!15}\footnotesize 
Counting & FSC147~\cite{m_Ranjan-etal-CVPR21}, TallyQA~\cite{acharya2019tallyqa} \\
\end{tabular}}
\label{tab:data_sft}
\end{table*}

\subsubsection{LocateAnything Fine-Tuning (Detection and Grounding Enhancement)}
Following the initial world-knowledge alignment, we then train the LocateAnything model using a carefully designed two-stage SFT strategy tailored for fine-grained detection and grounding. This constitutes the final two stages of our pipeline (leveraging the data presented in Fig. 5 of the main text). 
(1) \textbf{Stage 3 (Comprehensive Detection and Grounding):} We incorporate a massive mixture of 138M queries into the overall training data to equip the model with comprehensive grounding and detection capabilities. During this stage, all model components are fully unfrozen and trained. We set the maximum sequence length to 25,600 and employ a learning rate of $4 \times 10^{-5}$ with a Cosine schedule. 
(2) \textbf{Stage 4 (Dense Detection Enhancement):} To further boost the model's recall in dense scenes, we reduce the proportion of general training data to 20\% while significantly increasing the proportion of data containing many objects per image (\eg, MOT20Det, SKU110K). All components remain trainable, and the maximum sequence length is maintained at 25,600. The learning rate is decayed to $1 \times 10^{-5}$.

\subsubsection{Stream Packing}
A key challenge for training with our dual-formulation (NTP + MTP) design is that different samples, after block-wise expansion, exhibit highly variable sequence lengths. Na\"ive padding-based batching leads to significant GPU memory waste and low arithmetic utilization. To address this, we adopt an \textit{online stream packing} strategy that dynamically assembles multiple variable-length samples into a single, densely packed sequence of a target budget (\eg, 36,864 tokens). Concretely, our packing pipeline operates through three core mechanisms. First, via \textbf{Weighted Sampling}, an infinite iterator draws samples from multiple heterogeneous datasets according to pre-specified mixing weights. Second, utilizing \textbf{Best-Fit Buffering}, a fixed-size buffer (default size 32) stores pending samples. When assembling a batch, the packer first scans the buffer for the \textit{largest} sample that still fits into the remaining token budget---a best-fit decreasing heuristic that empirically yields $>$95\% packing efficiency. If no buffered sample fits, a freshly drawn sample is either appended directly (if it fits) or placed into the buffer for future use. Third, through \textbf{Big-Rocks-First Seeding}, after yielding a completed batch, the packer seeds the next batch with the \textit{largest} sample currently in the buffer, ensuring that oversized samples are never starved. Each packed sequence carries a \texttt{sub\_sample\_lengths} tensor that records the constituent sample boundaries. This metadata is consumed downstream by the attention kernel to construct the correct per-sample attention mask within the packed sequence, preventing cross-contamination between unrelated samples.

\subsubsection{MagiAttention for Heterogeneous Mask Training}
Our dual-formulation training produces a \textit{heterogeneous} attention mask that combines standard causal attention (for the NTP stream) with block-causal and bidirectional intra-block patterns (for the MTP stream), all within a single packed sequence potentially containing multiple samples. When further combined with stream packing, the resulting attention mask becomes highly irregular and sample-dependent, making it incompatible with conventional Flash-Attention kernels that assume a uniform causal or full-attention pattern.

To efficiently handle this, we leverage \textbf{MagiAttention}~\citep{magiattention2025}, a distributed attention framework designed for ultra-long contexts with heterogeneous masks. Together, stream packing and MagiAttention form a synergistic training infrastructure: packing maximizes token-level utilization within each GPU, while MagiAttention ensures that the resulting heterogeneous attention masks are handled both correctly and efficiently across the distributed training cluster.

\subsection{Inference Details}
We provide a detailed description of the inference pipeline, including the generation modes, the semi-autoregressive generation loop, the box-aware decoding strategies, and the hyperparameter configurations used across all evaluations.

\subsubsection{Generation Hyperparameters}
We employ nucleus sampling with a temperature of $0.7$ and top-$p$ of $0.9$ to balance diversity and precision. A repetition penalty of $1.1$ is applied to discourage duplicate predictions. KV cache is enabled throughout inference to avoid redundant computation. The block size for MTP generation is set to $6$ (\ie, $n_{\text{future}} = 6$), meaning each parallel decoding step predicts up to $6$ tokens simultaneously. The maximum number of newly generated tokens is set to $8{,}192$. All models are evaluated in BF16 precision with a batch size of $1$.

\subsubsection{KV Cache Management} 
After each MTP step, the KV cache is truncated to include only the positions corresponding to actually committed tokens (\ie, the prefix up to the current generation frontier). The mask tokens and the duplicated anchor token are evicted, ensuring that subsequent steps attend only to the ground-truth generation history. This truncation is essential for maintaining consistency between the causal prefix seen during training and the KV cache state during inference.

\section{LocateAnything-Data Construction}

\subsection{Leveraging Existing Open-Source Data}
We begin by collecting high-quality detection and grounding datasets from the open-source community and performing unified format cleaning and normalization. As illustrated in Fig.~\ref{fig:dataset_stats} of the main paper, the collected data span six domains, covering diverse visual scenarios.

Except for GroundCUA~\citep{feizi2025grounding}, we use the original labels for all other GUI datasets. The GroundCUA dataset, however, requires additional processing because its original labels typically correspond only to short descriptions of UI elements. To enrich the grounding queries for this specific dataset, we augment the GroundCUA annotations using Qwen3-VL~\citep{bai2025qwen3vltechnicalreport}. Specifically, given the original \textit{bbox}, \textit{label}, and \textit{category}, we first render the target bounding box on the screenshot and crop a local region around it. The full screenshot, the cropped region, together with the label, category, and platform metadata are then provided to Qwen3-VL~\citep{bai2025qwen3vltechnicalreport}. After determining whether the target element is visually identifiable, the model generates natural language descriptions from three complementary perspectives: \textit{appearance}, describing visual attributes such as color, shape, iconography, or textual content; \textit{spatial}, describing the element's relative position with respect to other UI components; and \textit{functional}, describing the user intent or interaction semantics associated with the element. Through this process, the original discrete text labels of GroundCUA are transformed into richer, multi-dimensional grounding queries that are both descriptive and interpretable.

Referring and grounding datasets themselves are relatively limited in scale. To address this, we aggregate several widely used benchmarks, including Flickr30k Entities~\citep{plummer2016flickr30kentitiescollectingregiontophrase}, gRefCOCO~\citep{he2023grecgeneralizedreferringexpression}, RefCOCO~\citep{yu2016modelingcontextreferringexpressions}, HumanPart~\citep{yu2016modelingcontextreferringexpressions}, and HumanRef~\citep{jiang2025rexomni}. In addition, we incorporate large-scale detection datasets such as OpenImages~\citep{kuznetsova2020open}, Objects365~\citep{Shao_2019_ICCV}, and images collected from Unsplash. These datasets serve as raw sources for constructing our multi-target grounding data engine, as discussed in Sec.~\ref{sec:Multi-Targets Grounding Data Engine}.

\begin{figure*}[!htbp]
    \centering
    \includegraphics[width=\textwidth]{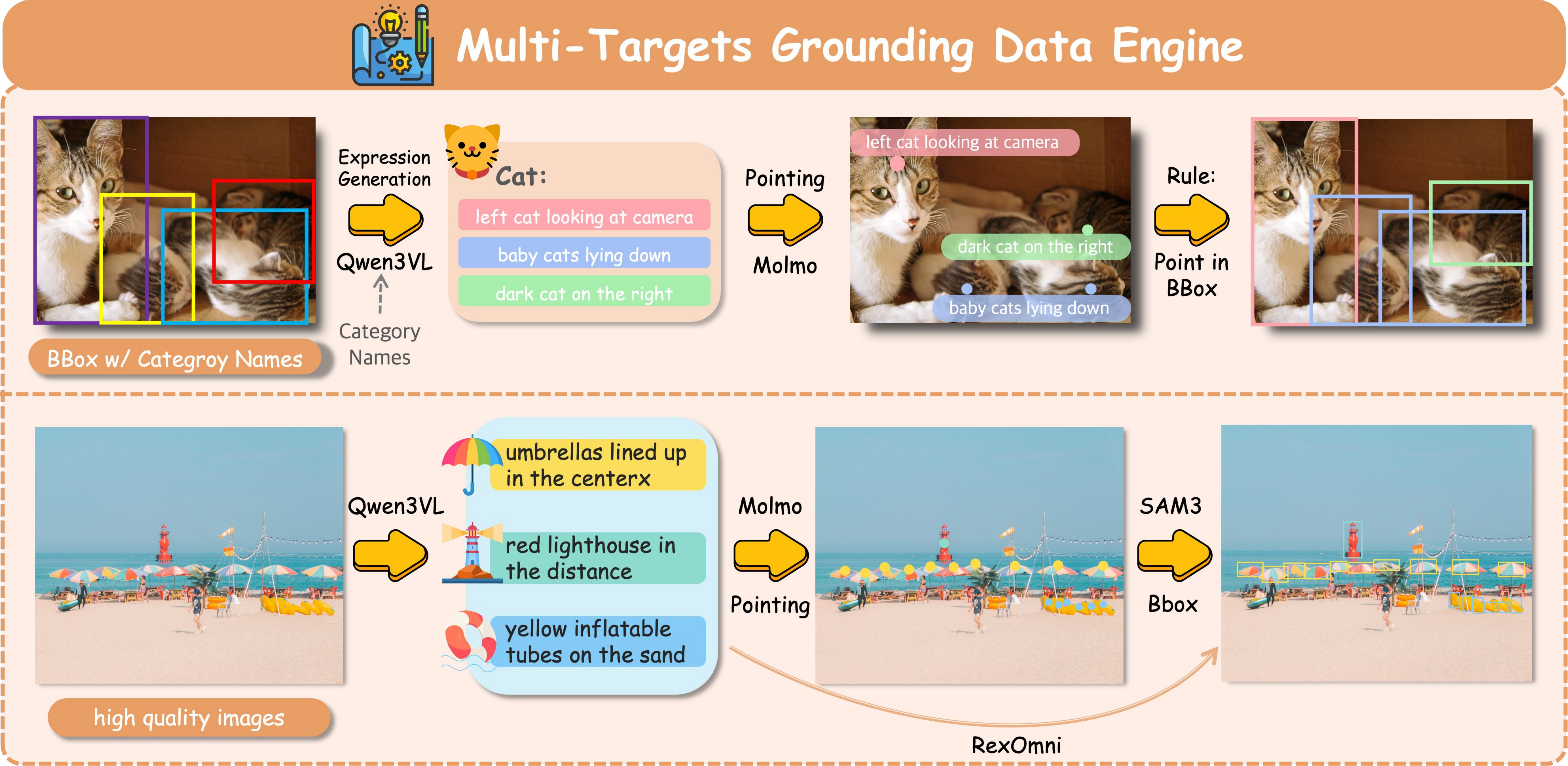}
    \caption{\textbf{Data engine for multi-targets grounding.} \textbf{Top:} For detection datasets with gt boxes, we use each box category as a prompt to Qwen3-VL~\citep{bai2025qwen3vltechnicalreport} to synthesize detailed object-centric queries, including attributes, spatial relations, and reasoning cues. These queries are then fed to Molmo~\citep{deitke2025molmo} to predict candidate points, from which we retain the points falling inside the corresponding gt boxes as reliable supervision. \textbf{Bottom:} For a large collection of high-quality unlabeled images, Qwen3-VL directly generates diverse queries from the image. Such queries can be used to prompt Molmo for point prediction, followed by SAM 3~\citep{carion2025sam3segmentconcepts} to produce boxes, or directly prompt Rex-Omni~\citep{jiang2025rexomni} to generate boxes. All generated boxes are finally post-verified by Qwen3-VL.}
    \label{fig:multi-targets-data-engine}
    \vspace{1mm}
\end{figure*}

Another critical issue in existing detection and grounding datasets is that they almost exclusively contain \textit{positive} samples. Training on such data can lead to hallucination behaviors, where the model predicts bounding boxes even when the query is unrelated to the image. To mitigate this issue, we explicitly construct negative samples across domains. The proportion of negative queries varies depending on the domain statistics (see Tab.~\ref{tab:data_statistics}). Concretely, we generate queries referring to objects that do not exist in the image, and assign them the \textit{negative block} described in Fig.~\ref{fig:method1} of the main paper. This design enables the model to learn to abstain when no valid grounding target is present.

\subsection{Multi-Targets Grounding Data Engine}
\label{sec:Multi-Targets Grounding Data Engine}

Existing open-source grounding datasets are relatively limited in scale and diversity. To construct a large-scale multi-target grounding dataset, we design a data engine that automatically synthesizes grounding annotations from both labeled detection data and unlabeled images, as illustrated in Fig.~\ref{fig:multi-targets-data-engine}.

\paragraph{From Detection Datasets:}
We first leverage high-quality detection datasets such as Open Images~\citep{kuznetsova2020open} and Objects365~\citep{Shao_2019_ICCV}. For each ground-truth bounding box, we use its category label as a prompt to Qwen3-VL~\citep{bai2025qwen3vltechnicalreport} to generate a set of detailed object-centric queries, including attributes, spatial relations, and reasoning cues. These queries are then used to prompt Molmo~\citep{deitke2025molmo} to predict candidate points. Since the ground-truth boxes are known, we retain only the points that fall inside the corresponding bounding boxes, which serve as reliable grounding supervision.

\paragraph{From Unlabeled Images:}
To further expand the diversity of grounding targets, we additionally collect large amounts of high-quality unlabeled images from Unsplash and SA-1B~\citep{kirillov2023segment}. For each image, Qwen3-VL directly generates a diverse set of natural language queries describing potential objects or regions. These queries can be used to prompt Molmo~\citep{deitke2025molmo} to predict points, which are subsequently converted into bounding boxes using SAM 3~\citep{carion2025sam3segmentconcepts}. Alternatively, the queries can directly prompt Rex-Omni~\citep{jiang2025rexomni} to predict bounding boxes. 

To ensure annotation quality, all generated boxes are finally verified by Qwen3-VL\citep{bai2025qwen3vltechnicalreport} through a post-checking stage, filtering out inconsistent predictions.

\subsection{Task-Specific Prompt Design}

As detailed in Tab.~\ref{tab:task_prompts}, we present a comprehensive overview of the versatile perception tasks supported by our unified framework, alongside their corresponding output formats and question templates. To seamlessly integrate diverse visual grounding and detection capabilities, we design specific textual prompts for each task. The model handles a wide spectrum of region-based tasks that output bounding boxes, including Object Detection, Text Grounding, Scene Text Detection, and Document Layout Analysis. Furthermore, it supports fine-grained localization tasks such as Pointing, which outputs specific coordinate points. For complex referring and interactive tasks like Phrase Grounding and GUI Grounding, the framework flexibly predicts either single/multiple boxes or points depending on the user's intent. Within the prompt templates, \texttt{[PHRASE]} represents a free-form natural language description, while \texttt{[CATEGORIES]} denotes a comma-separated list of target category names. This unified prompting strategy enables the model to effectively bridge natural language instructions with precise spatial coordinate decoding.

\definecolor{nvgreen}{RGB}{118, 185, 0}
\definecolor{nvheader}{RGB}{118, 185, 0}
\definecolor{nvlightgreen}{RGB}{240, 249, 226}
\definecolor{nvwhite}{RGB}{255, 255, 255}
\definecolor{nvfont}{RGB}{255, 255, 255}

\begin{table}[H]
\centering
\vspace{+2mm}
\caption{Overview of supported perception tasks and their corresponding prompt templates. \texttt{[PHRASE]} denotes a free-form natural language description, and \texttt{[CATEGORIES]} denotes a comma-separated list of category names.}
\label{tab:task_prompts}
\renewcommand{\arraystretch}{1.3}
\setlength{\tabcolsep}{6pt}
\small

\renewcommand{\tabularxcolumn}[1]{m{#1}}
\begin{tabularx}{\linewidth}{>{\raggedright\arraybackslash}m{3.2cm} c >{\raggedright\arraybackslash}X}
\arrayrulecolor{nvgreen}
\toprule
\rowcolor{nvheader}
\textcolor{nvwhite}{\textbf{Task}} & \textcolor{nvwhite}{\textbf{Output}} & \textcolor{nvwhite}{\textbf{Question Template}} \\
\midrule

\rowcolor{nvlightgreen}
\textbf{Object Detection} & Box & Locate all the instances that matches the following description:  \texttt{[CATEGORIES]}. \\
\midrule\noalign{\vspace{3pt}}

\rowcolor{nvwhite}
 & Single Box & Locate a single instance that matches the following description: \texttt{[PHRASE]}. \\

\rowcolor{nvwhite}
\multirow{-2}{3.2cm}{\textbf{Phrase Grounding}} & Multiple Boxes & Locate all the instances that match the following description: \texttt{[PHRASE]}. \\
\midrule

\rowcolor{nvlightgreen}
\textbf{Text Grounding} & Box & Please locate the text referred as \texttt{[PHRASE]}. \\
\midrule

\rowcolor{nvwhite}
\textbf{Scene Text Detection} & Box & Detect all the text in box format. \\
\midrule

\rowcolor{nvlightgreen}
\textbf{Document Layout Analysis} & Box & Detect all the objects in the image that belong to the category set: \texttt{[CATEGORIES]}. \\
\midrule\noalign{\vspace{3pt}}

\rowcolor{nvwhite}
 & Box & Locate the region that matches the following description: \texttt{[PHRASE]}. \\
\rowcolor{nvwhite}
\multirow{-2}{3.2cm}{\textbf{GUI Grounding}} & Point & Point to: \texttt{[PHRASE]}. \\
\midrule

\rowcolor{nvlightgreen}
\textbf{Pointing} & Point & Point to: \texttt{[PHRASE]}. \\
\bottomrule

\end{tabularx}
\end{table}

\subsection{Data Statistics and Distribution}
We analyze the statistical characteristics of the collected dataset.
Tab.~\ref{tab:data_statistics} summarizes the dataset statistics across six domains. 
In total, the dataset contains over 139M queries with more than 22M negative samples. 

\begin{figure*}[t]
    \centering
    \includegraphics[width=\textwidth]{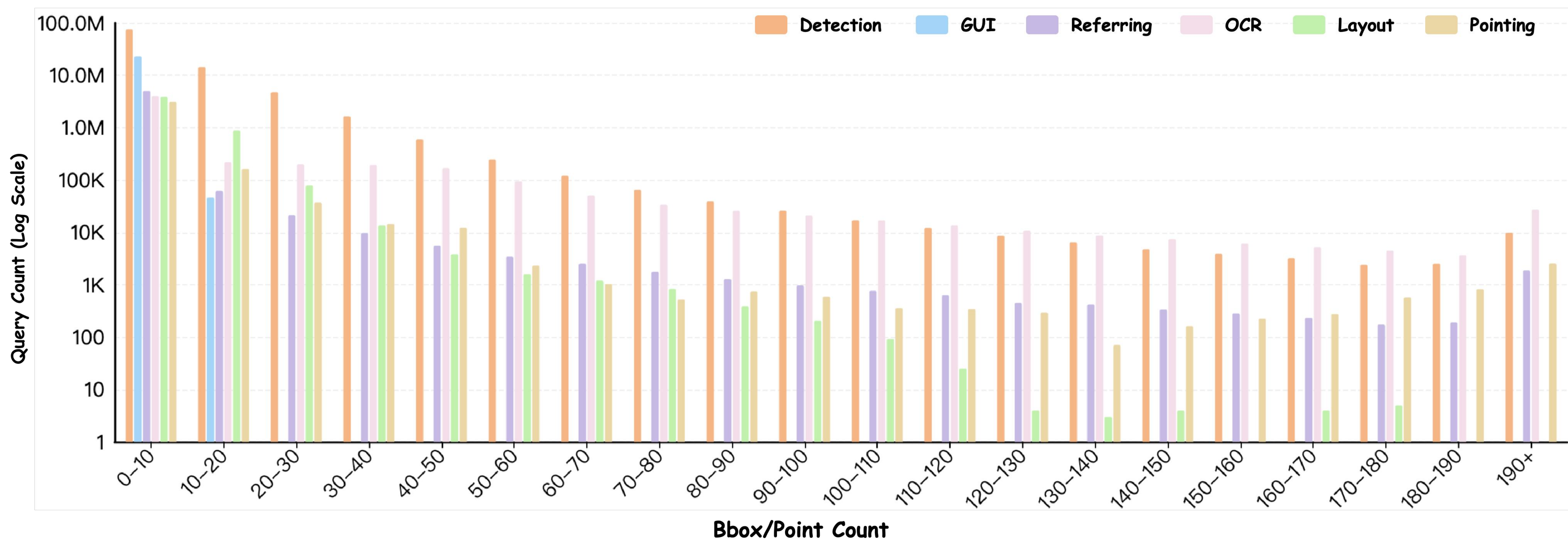}
    \caption{Distribution of the number of targets per query across different domains. The x-axis shows the number of targets associated with a query, while the y-axis (log scale) indicates the number of queries. }
    \label{fig:target-per-query}
\end{figure*}

Our dataset also exhibits strong multi-target grounding characteristics. 
The number of targets associated with each query varies substantially across domains. 
As illustrated in Fig.~\ref{fig:target-per-query}, the distribution of targets per query follows a long-tailed pattern: most queries correspond to a small number of targets, while a non-negligible portion involve a large number of instances.

\begin{figure*}[!htbp]
    \vspace{+2mm}
    \centering
    \includegraphics[width=\textwidth]{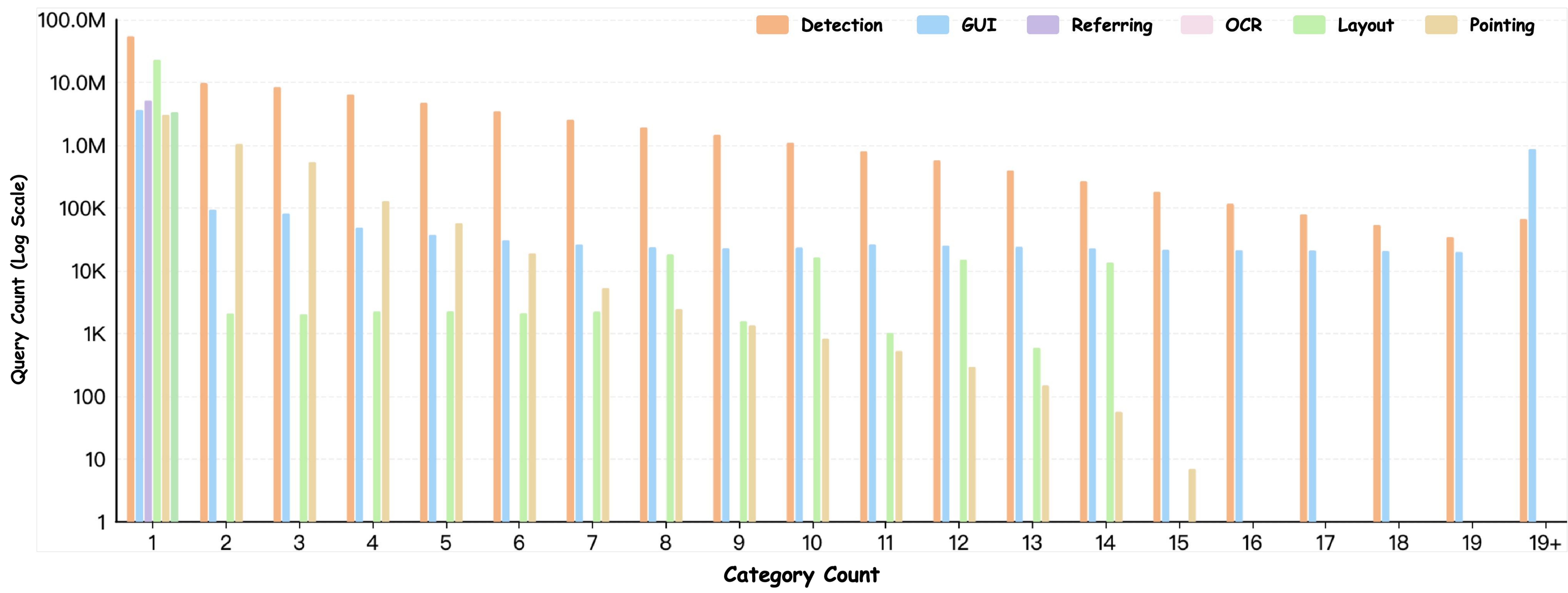}
    \caption{Distribution of query length across domains. The x-axis represents the number of words in the target description (excluding template text), and the y-axis (log scale) shows the number of queries.}
    \label{fig:categroy-per-query}
\end{figure*}

We further analyze the linguistic properties of the queries. 
As shown in Fig.~\ref{fig:categroy-per-query}, query length varies across domains, reflecting different grounding paradigms and language patterns used to describe visual targets.

Overall, these statistics highlight the scale, diversity, and multi-target nature of our dataset, which together provide a strong foundation for training models capable of handling heterogeneous visual domains and complex language queries.
\begin{table*}[!t]
\centering
\caption{Statistics of the collected data across six domains. We report the total number of queries and negative samples, together with the maximum and mean numbers of targets and categories per query (/ Q), and targets per image (/ I). \textit{Query length} measures the number of words in the target description after removing template text, reflecting the actual linguistic content used to describe grounding targets.}
\vspace{-1mm}
\label{tab:data_statistics}
\resizebox{1.0\textwidth}{!}{
\setlength{\tabcolsep}{6pt}
\renewcommand{\arraystretch}{1.2}
\begin{tabular}{l|c|c|cc|cc|cc|cc}
\toprule
\multirow{2}{*}{\textbf{Domain}} &
\multirow{2}{*}{\textbf{\#Queries}} &
\multirow{2}{*}{\textbf{\#Negative}} &
\multicolumn{2}{c|}{\textbf{Targets / Q}} &
\multicolumn{2}{c|}{\textbf{Categories / Q}} &
\multicolumn{2}{c|}{\textbf{Query Length}} &
\multicolumn{2}{c}{\textbf{Targets / I}} \\
\cline{4-5}\cline{6-7}\cline{8-9}\cline{10-11}
& & &
\multicolumn{1}{c}{\textbf{Max}} &
\multicolumn{1}{c}{\textbf{Mean}} &
\multicolumn{1}{c}{\textbf{Max}} &
\multicolumn{1}{c}{\textbf{Mean}} &
\multicolumn{1}{c}{\textbf{Max}} &
\multicolumn{1}{c}{\textbf{Mean}} &
\multicolumn{1}{c}{\textbf{Max}} &
\multicolumn{1}{c}{\textbf{Mean}} \\
\midrule
Detection & 93,351,373 & 21,021,509 & 745  & 6.29  & 43   & 2.47 & 251 & 24.19 & 3,725 & 30.68 \\
GUI       & 23,009,535 & 0          & 14   & 1.03  & 14   & 1.03 & 351 & 4.07  & 8,690 & 7.95  \\
Referring & 10,141,597 & 93,396     & 818  & 2.12  & 1    & 0.89 & 53  & 5.48  & 6,938 & 9.65  \\
OCR       & 5,052,040  & 0          & 2,337 & 11.89 & 1,258 & 10.4 & 51  & 1.17  & 2,337 & 28.67 \\
Layout    & 4,859,914  & 1,384,804  & 176  & 4.92  & 15   & 1.31 & 30  & 2.2   & 880   & 21.17 \\
Pointing  & 3,148,098  & 353,366    & 675  & 3.25  & 1    & 0.89 & 189 & 2.63  & 1,575 & 14.92 \\
\bottomrule
\end{tabular}
}
\vspace{-1mm}
\end{table*}

\section{Additional Experiments}

\subsection{Results on Pointing Tasks}
To further evaluate the fine-grained spatial perception of our model, we benchmark LocateAnything-3B on point-based localization tasks, where the model must predict a point that falls within the target's bounding box or segmentation mask. As detailed in Tab.~\ref{tab:pointing}, LocateAnything-3B (evaluated under Hybrid Mode) achieves state-of-the-art results across a diverse suite of benchmarks.

It significantly outperforms contemporary vision-language models, including larger networks like OVIS2.5-9B and point-centric specialists such as Rex-Omni-3B. Notably, our model scores 83.9 F1@Point on COCO and exhibits exceptional resilience in heavily packed environments, reaching 87.6 F1@Point on Dense200. Furthermore, it demonstrates superior alignment of complex human intents to spatial regions, achieving 84.7 F1@Point on HumanRef and 91.0 F1@Point on the RefCOCOg test set. These results underscore the effectiveness of our box-aligned training paradigm and the massive scale of LocateAnything-Data in establishing precise geometric alignments, extending seamlessly to point-based generation.

\begin{table*}[!t]
\vspace{+5mm}
\caption{Performance evaluation for the object pointing task across a diverse range of benchmarks (COCO, LVIS, Dense200, VisDrone, RefCOCOg, HumanRef). F1-scores are used as the primary metric. The results of the \textit{Hybrid Mode} are reported here.}
    \centering
    \resizebox{1.0\textwidth}{!}{
       \begin{tabular}{c|cc|cc|ccc}
\toprule
                         & COCO                              & LVIS                              & Dense200                    & VisDrone      & HumanRef      & RefCOCOg val  & RefCOCOg test \\ \cline{2-8} 
\multirow{-2}{*}{Method} & F1@Point                          & F1@Point                          & F1@Point                    & F1@Point      & F1@Point      & F1@Point      & F1@Point      \\ \midrule
OVIS2.5-2B               & 73.4                              & 52.8                              & 36.4                        & 23.8          & 72.5          & 83.1          & 83.1          \\
Qwen2.5-VL-3B            & 65.9                              & 48.3                              & 4.3                         & 13.9          & 64.1          & 77.4          & 77.8          \\
Qwen2.5-VL-7B            & 61.1                              & 56.5                              & 2.0                         & 14.2          & 65.1          & 78.9          & 79.4          \\
OVIS2.5-9B               & 72.6                              & 61.7                              & 35.0                        & 18.8          & 62.3          & \underline{85.0}& 84.5          \\
Molmo-7B-D               & 77.3                              & 40.3                              & 33.1                        & 29.2          & 70.0          & 83.7          & 83.6          \\
SEED1.5-VL               & {\color[HTML]{000000} 78.2}       & {\color[HTML]{000000} 70.7}       & {\color[HTML]{000000} 72.1} & 56.7          & 83.1          & 83.6          & 84.2          \\
Rex-Omni-SFT-3B          & 76.0                              & 66.7                              & 72.9                        & 49.5          & 82.1          & 83.3          & 83.9          \\
Rex-Omni-3B              & \underline{80.5}                  & \underline{70.8}                  & \underline{82.5}            & \underline{58.9}& \underline{83.8}& 84.7          & \underline{85.1}\\ 
\rowcolor{lightblue!10} LocateAnything-3B                & \textbf{83.9}                     & \textbf{76.6}                     & \textbf{87.6}               & \textbf{60.4} & \textbf{84.7} & \textbf{91.3} & \textbf{91.0} \\ \bottomrule
\end{tabular}
    }
    \centering
    \label{tab:pointing}
    \vspace{-1em}
\end{table*}

\subsection{Comprehensive Performance Across Decoding Modes}
In this section, we provide a detailed breakdown of LocateAnything's performance across its three on-demand decoding modes: \textbf{Fast}, \textbf{Hybrid}, and \textbf{Slow}. These modes allow for a dynamic trade-off between geometric precision and inference latency, as summarized in Tab.~\ref{tab:grand_summary_tasks}.

\begin{table*}[!t]
    \centering
    \vspace{+7mm}
    \caption{Comprehensive performance of our Fast, Hybrid, and Slow configurations across multiple visual tasks. Throughput (measured in Boxes Per Second, BPS) is reported in the header for each mode. For general detection (COCO, LVIS), we report Average Precision (AP), Average Recall (AR), and F1@mIoU. For other tasks, we report the primary comprehensive metric (e.g., F1@mIoU, F1@mIoU, Avg Acc).}
    \label{tab:grand_summary_tasks}
    \resizebox{1.0\textwidth}{!}{
        \renewcommand{\arraystretch}{1.2}
        \setlength{\tabcolsep}{10pt}
        \begin{tabular}{ll c | c | c | c}
            \toprule
            \multirow{2}{*}{\textbf{Task Group}} & \multirow{2}{*}{\textbf{Dataset}} & \multirow{2}{*}{\textbf{Metric}} & \textbf{Fast Model} & \textbf{Hybrid Model} & \textbf{Slow Model} \\
            & & & (15.3 BPS) & (12.7 BPS) & (4.3 BPS) \\
            \midrule
            
            \multirow{8}{*}{\textbf{Detection}} 
            & \multirow{3}{*}{COCO} 
            & P@mIoU & \underline{58.9} & \textbf{60.8} & \textbf{60.8} \\
            & & R@mIoU & 46.8 & \underline{49.7}  & \textbf{50.3} \\
            & & F1@mIoU & 52.2 & \underline{54.7} & \textbf{55.1} \\
            \cline{2-6} \rule{0pt}{10pt}
            & \multirow{3}{*}{LVIS} 
            & P@mIoU & 64.3 & \textbf{68.4} & \underline{68.0} \\
            & & R@mIoU & 37.1 & \underline{40.3} & \textbf{42.8} \\
            & & F1@mIoU & 47.0 & \underline{50.7} & \textbf{52.6} \\
            \cline{2-6} \rule{0pt}{10pt}
            & Dense200 & F1@mIoU & 46.8 & \underline{61.3} & \textbf{61.5} \\
            & VisDrone & F1@mIoU & 34.4 & \underline{39.8} & \textbf{40.2} \\
            \midrule
            
            \multirow{4}{*}{\textbf{OCR}} 
            & HierText & F1@mIoU & 28.8 & \underline{29.1} & \textbf{43.2} \\
            & ICDAR2015 & F1@mIoU & 26.6 & \underline{26.4} & \textbf{27.3} \\
            & TotalText & F1@mIoU & 44.4 & \underline{44.6} & \textbf{47.5} \\
            & SROIE & F1@mIoU & 38.8 & \underline{39.3} & \textbf{64.4} \\ 
            \midrule
            
            \multirow{2}{*}{\textbf{Layout}} 
            & DocLayNet & F1@mIoU & 67.2 & \underline{77.7} & \textbf{80.4} \\
            & M6Doc & F1@mIoU & 64.1 & \textbf{70.5} & \underline{69.7} \\
            \midrule

            \textbf{GUI} 
            & ScreenSpot-Pro & Acc & 59.7 & \underline{60.3} & \textbf{60.5} \\
            \midrule

            \multirow{3}{*}{\textbf{Referring}} 
            & HumanRef & F1@mIoU & 66.8 & \underline{78.5} & \textbf{79.1} \\
            & RefCOCOg val & F1@mIoU & 70.8 & \textbf{73.4} & \underline{72.4} \\
            & RefCOCOg test & F1@mIoU & 72.5 & \textbf{74.8} & \underline{73.8} \\
            \midrule

            \multirow{4}{*}{\textbf{Pointing}} 
            & COCO & F1@Point & 83.1 & \underline{83.9} & \textbf{84.8} \\
            & LVIS & F1@Point & 74.4 & \underline{76.6} & \textbf{76.9} \\
            & Dense200 & F1@Point & \textbf{89.4} & 87.6 & \underline{88.3} \\
            & VisDrone & F1@Point & 58.1 & \underline{60.4} & \textbf{61.3} \\

            \bottomrule
        \end{tabular}
    }
\vspace{-2mm}
\end{table*}

\subsection{Backbone Generalization}
To examine whether Parallel Box Decoding (PBD) depends on a specific vision-language backbone, we instantiate the same decoding design on Qwen3-VL-4B~\citep{bai2025qwen3vltechnicalreport}. Following the controlled setting used for the ablation study in the main paper, this variant is trained exclusively on COCO, isolating the effect of the decoding formulation from large-scale data scaling.

\begin{wraptable}[6]{r}{0.40\textwidth}
    \vspace{-5mm}
    \captionsetup{
        width=\linewidth,
        justification=raggedright,
        singlelinecheck=false
    }
    \setlength{\abovecaptionskip}{0pt}
    \setlength{\belowcaptionskip}{2pt}
    \caption{\textbf{Backbone generalization.}}
    \label{tab:backbone_generalization}
    \small
    \renewcommand{\arraystretch}{0.88}
    \setlength{\tabcolsep}{3.2pt}
    \resizebox{\linewidth}{!}{%
        \begin{tabular}{@{}llcc@{}}
            \toprule
            \textbf{Backbone} & \textbf{Decoding} & \textbf{F1} & \textbf{BPS} \\
            \midrule
            Qwen3-VL-4B & NTP (baseline) & 50.8 & 2.8 \\
            Qwen3-VL-4B & + PBD (Slow) & \textbf{52.2} & 2.8 \\
            Qwen3-VL-4B & + PBD (Fast) & 49.6 & \textbf{11.4} \\
            \rowcolor{lightblue!10}
            Qwen3-VL-4B & + PBD (Hybrid) & \underline{52.0} & \underline{9.4} \\
            \bottomrule
        \end{tabular}%
    }
    \vspace{-5mm}
\end{wraptable}

As shown in Tab.~\ref{tab:backbone_generalization}, PBD consistently improves the speed--accuracy trade-off on Qwen3-VL-4B. The Hybrid configuration improves COCO F1 from $50.8$ to $52.0$ while increasing throughput from $2.8$ to $9.4$ BPS. These results indicate that the benefits of box-aligned parallel decoding are not tied to a particular backbone architecture.

\subsection{Mode Analysis and Throughput}
Our on-demand decoding modes allow for a dynamic trade-off between geometric precision and inference latency. (1) \textbf{Slow Mode (Next-Token Prediction):} Utilizing standard autoregressive generation, this mode consistently establishes the upper bound for localization accuracy (\eg, peak F1@mIoU of 55.1 on COCO and 79.8 on DocLayNet). By processing tokens sequentially, it maintains superior spatial awareness and robust handling of dense object clusters. (2) \textbf{Fast Mode (Multi-Token Prediction):} This mode maximizes inference throughput to \textbf{15.3 BPS} by predicting full geometric elements in parallel. While it incurs slight accuracy drops in complex or highly dense scenarios, its high-velocity output makes it ideal for real-time applications. (3) \textbf{Hybrid Mode (Adaptive Decoding):} Serving as the optimal choice for production pipelines, this mode defaults to parallel decoding and selectively falls back to autoregressive generation only when spatial ambiguity or format irregularities are detected. Operating at a highly competitive \textbf{12.7 BPS}, it preserves the speed gains of parallelization while maintaining precise outputs.

\subsection{Experimental Setup}
To ensure transparency and reproducibility, all performance metrics are reported under specific configurations. For \textbf{Throughput Benchmarking}, all values, measured in Boxes Per Second (BPS), were evaluated specifically on the COCO dataset to provide a consistent baseline for speed comparison. Regarding \textbf{Input Resolution}, images for the \textbf{COCO} and \textbf{LVIS} benchmarks were resized with the short side set to 840 pixels. For all other benchmarks, the model was evaluated using the original resolution of the source data.

\clearpage

\subsection{Qualitative Comparisons}
\label{sec:qualitative_comparison}

\begin{figure*}[!htbp]
    \vspace{-1mm}
    \centering
    \includegraphics[width=\textwidth]{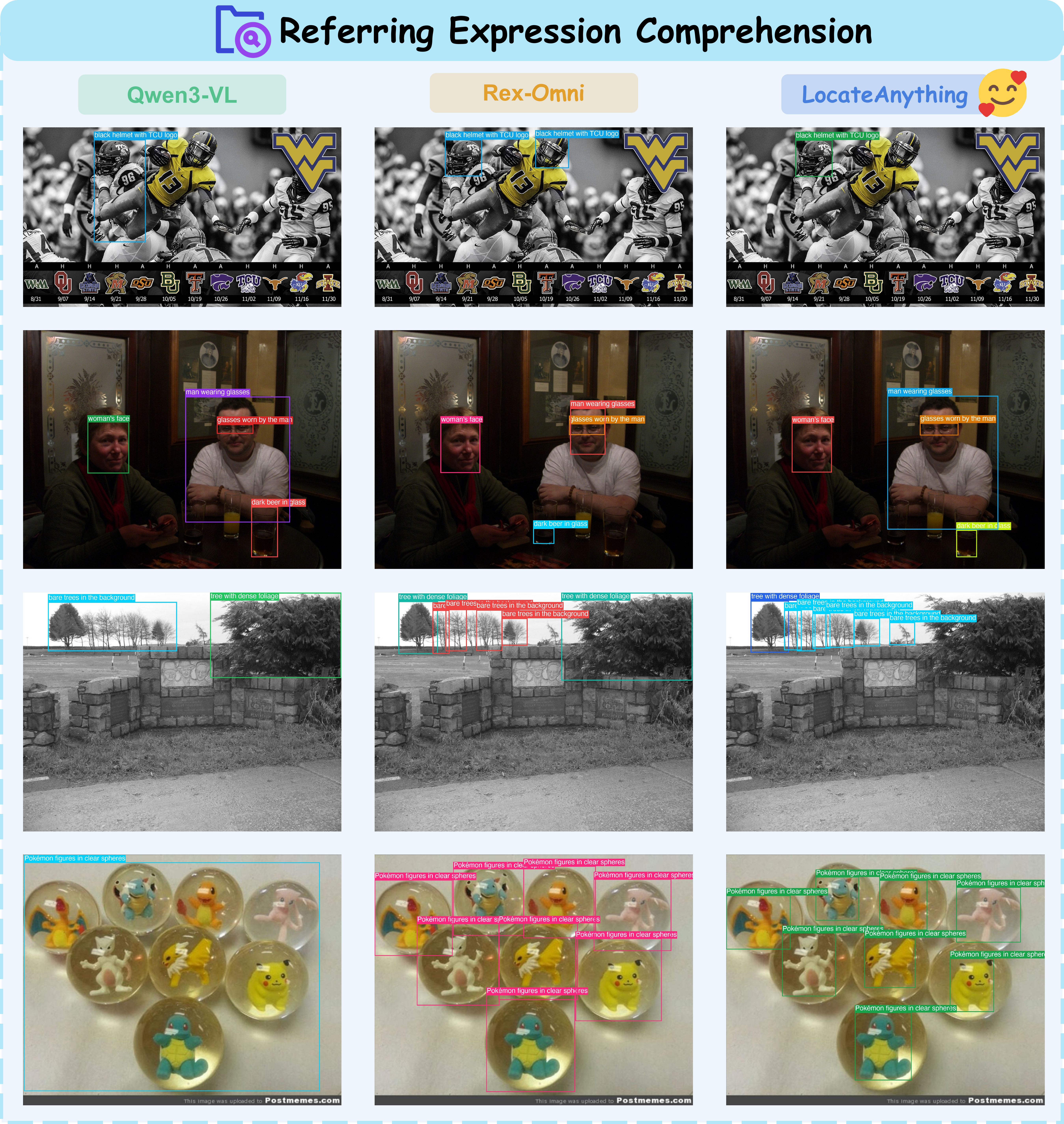}
    \caption{\textbf{Qualitative comparison on Referring Expression Comprehension (REC).} Compared to Qwen3-VL and Rex-Omni, LocateAnything demonstrates superior compositional grounding capabilities. It accurately aligns nuanced, free-form human intents (e.g., spatial or attribute-based queries) with correct visual regions.}
    \label{fig:vis_rec}
\end{figure*}

\begin{figure*}[!htbp]
    \centering
    \includegraphics[width=\textwidth]{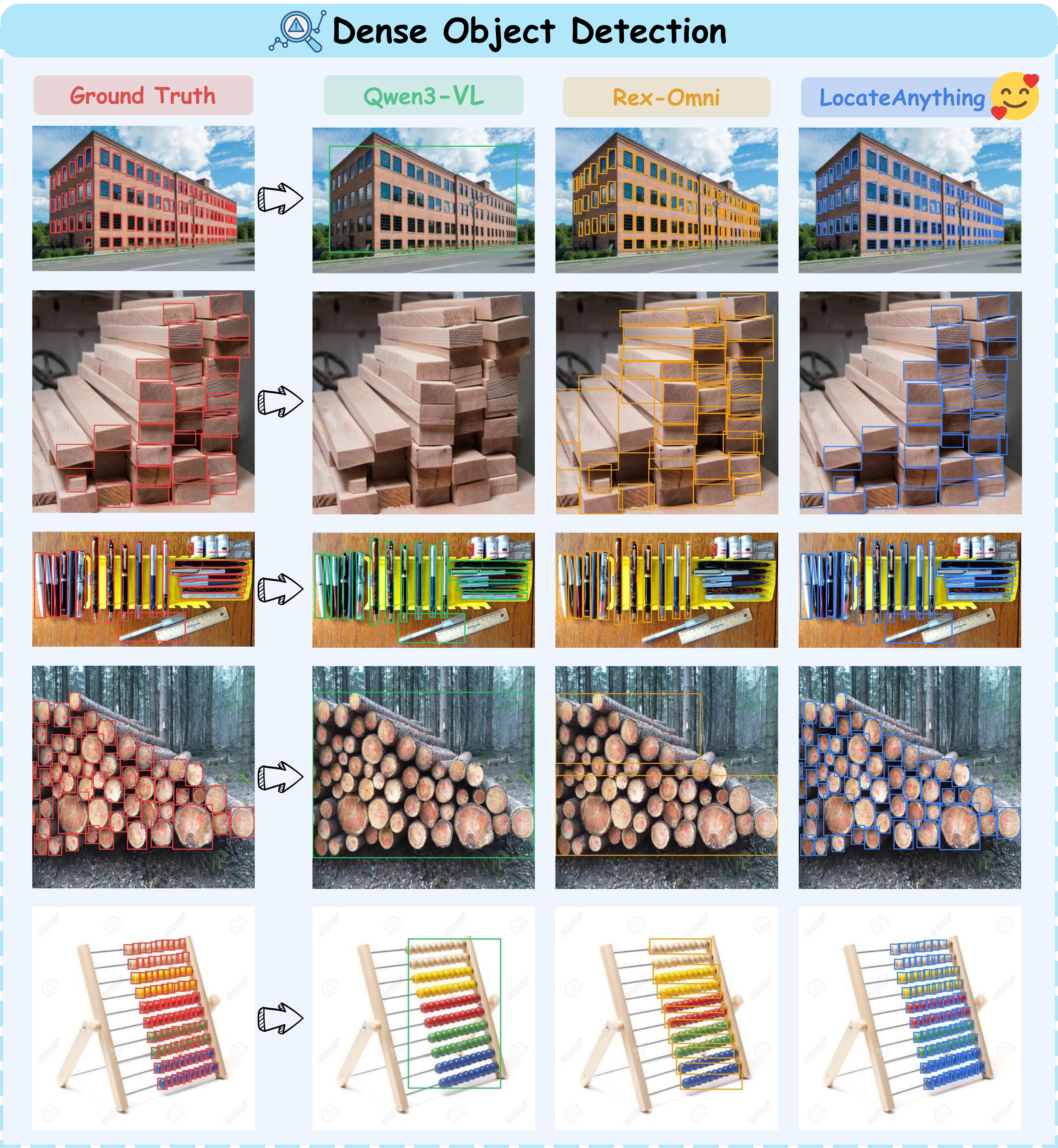}
    \caption{\textbf{Qualitative comparison on Dense Object Detection (DOD).} This figure illustrates performance in highly dense and heavily overlapping environments, such as stacked logs and abacus beads. While traditional token-by-token generation models (Qwen3-VL) and point-based models (Rex-Omni) suffer from severe omissions or spatial ambiguity (blurring boundaries between adjacent objects), LocateAnything maintains compact, well-separated, and highly accurate bounding boxes. This confirms the effectiveness of our block-level intra-attention and dense-aware Stage-2 training.}
    \label{fig:vis_dod}
\end{figure*}

\begin{figure*}[!htbp]
    \centering
    \includegraphics[width=\textwidth]{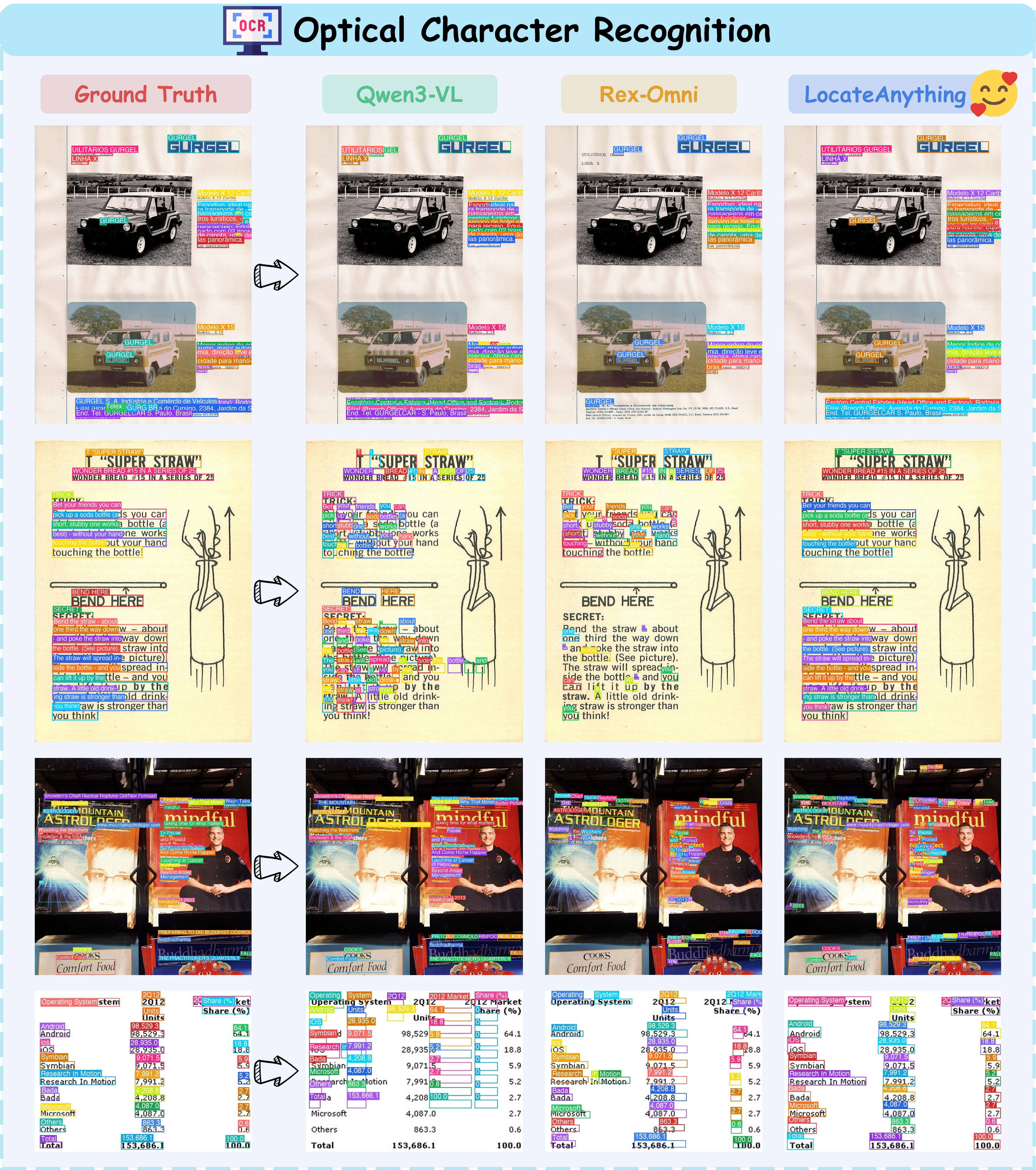}
    \caption{\textbf{Qualitative comparison on Optical Character Recognition (OCR).} For scene text (e.g., magazine covers) and structured documents (e.g., tables), LocateAnything yields tightly bounded boxes around text elements. The baseline models frequently exhibit format irregularities or merge distinct text blocks. Our parallel decoding, combined with the Hybrid Mode fallback for complex spatial layouts, ensures high-precision localization without sacrificing structural coherence.}
    \label{fig:vis_ocr}
\end{figure*}

\clearpage
\bibliographystyle{plainnat}
\bibliography{main}

\end{document}